\pdfoutput=1
\documentclass{article}

\PassOptionsToPackage{numbers,sort}{natbib}%

\usepackage[main, final]{neurips_2026}

\usepackage{graphicx}
\usepackage{subcaption}
\usepackage{float}
\usepackage[justification=justified]{caption}	%
\usepackage{lscape}                                         %
\usepackage{wrapfig}

\usepackage[lined,ruled,linesnumbered]{algorithm2e}
\usepackage{animate} %

\usepackage{booktabs}                   %
\usepackage{multirow}

\usepackage{makecell}

\usepackage{paralist}
\usepackage{enumitem}
\usepackage{pifont}
\usepackage{bm}                          %
\usepackage{epsfig}                      %
\usepackage{graphicx}                  %
\usepackage{times}
\usepackage{mathptmx}
\usepackage{mathtools}
\usepackage{amssymb,amsmath}   %
\usepackage{bbm}

\usepackage{units}
\usepackage{color}
\usepackage[T1]{fontenc}    %
\usepackage{amsfonts}       %
\usepackage[utf8]{inputenc} %
\usepackage{nicefrac}       %
\usepackage{comment}

\usepackage{url}  %
\usepackage[pagebackref=true,breaklinks=true,colorlinks,bookmarks=false]{hyperref}
\usepackage{xspace}
\usepackage[table]{xcolor}
\usepackage{setspace}
\usepackage{grfext}
\PrependGraphicsExtensions*{.jpg,.png,.PNG}
\setlist[itemize]{itemsep=2pt, topsep=2pt, parsep=1pt, partopsep=0pt}
\usepackage{wrapfig}

\def\eg{e.g.,~}               %
\def\ie{i.e.,~}               %

\newlength\paramargin
\newlength\figmargin

\newlength\secmargin
\newlength\figcapmargin
\newlength\tabcapmargin
\newlength\tabmargin

\newcommand{\figcaption}[2]
{
\caption{
\textbf{#1.}  %
#2            %
}
}

\newcommand{\secref}[1]{Section~\ref{sec:#1}}
\newcommand{\figref}[1]{Figure~\ref{fig:#1}} 
\newcommand{\tabref}[1]{Table~\ref{tab:#1}}
\newcommand{\eqnref}[1]{\eqref{eq:#1}}

\newcommand{\apref}[1]{Appendix~\ref{appen:#1}}

\long\def\ignorethis#1{}

\newbox\jsavebox%

\def\ours{{\texttt{TraCE}}}

\def\xi{\mathbf{x}_i}

\graphicspath{{figure}, {example}}

\title{TraCE: Trajectory-Aware Concepts\\
for Explainable Video Understanding}

\author{%
  \textbf{Wooil Lee}\textsuperscript{\normalfont 1}\thanks{Equal contribution.} \quad
  \textbf{Junpyo Hong}\textsuperscript{\normalfont 1}\footnotemark[1] \quad
  \textbf{Jongseo Lee}\textsuperscript{\normalfont 1} \\
  \textbf{Tuomas Oikarinen}\textsuperscript{\normalfont 2} \quad
  \textbf{Tsui-Wei Weng}\textsuperscript{\normalfont 2} \quad
  \textbf{Jinwoo Choi}\textsuperscript{\normalfont 1}\thanks{Corresponding author.} \\
  \textsuperscript{\normalfont 1}Kyung Hee University \qquad
  \textsuperscript{\normalfont 2}University of California San Diego \\
  \texttt{\{lwi2765,intpboy,jong980812,jinwoochoi\}@khu.ac.kr} \\
  \texttt{\{toikarinen,lweng\}@ucsd.edu}
}

\begin{document}
\maketitle

\begin{abstract}

Video understanding models often rely on motion, yet their explanations
can obscure which motion sources support a prediction.
We identify \emph{motion-source under-specification} as a limitation
of existing video XAI: region-centric explanations can be
source-ambiguous, while motion vocabularies restricted to single-person
pose sequences are source-incomplete.
We propose \ours{}---Trajectory-Aware Concepts for Explainable Video
Understanding---a concept-based framework that discovers
\emph{source-flexible} motion concepts by clustering optical-flow
trajectories from training videos.
These concepts represent dynamic evidence from people, objects, and
interactions, while object and scene concepts account for static context.
The resulting typed trajectory/object/scene interface supports
class prediction for action recognition and question-conditioned
concept matching for video question answering (VQA).
We evaluate \ours{} across five random seeds on six benchmarks:
NTU RGB+D Mutual, Chiral SSv2, UCF-101, KTH, TOMATO, and STAR.
Relative to non-interpretable baselines, \ours{} maintains competitive
task performance, with an average accuracy drop of 1.0 points for
action recognition and a nominal gain of 0.4 points for VQA.
Motion perturbation and concept-removal tests provide complementary
evidence of directional semantic alignment and decision faithfulness,
while sample-level interventions illustrate targeted model debugging.
Our project page is available at \url{https://leewooil.github.io/TraCE_project_page/}.

\end{abstract}

\section{Introduction}
\label{sec:intro}

\vspace{-0.1em}

Motion is a central source of evidence in video understanding.
Many events that share similar objects and scenes can differ primarily in how entities move over time~\cite{simonyan2014two,huang2018makes,only}. 
However, modern video understanding models~\cite{timesformer,videomae,videollava,wang2024internvideo2,qwen3vl} remain largely opaque:
their predictions do not reveal whether they rely on dynamic motion, static context, or spurious correlations~\cite{mall, rcore}.
Reliable deployment therefore requires explainable video understanding methods that expose the evidence behind each prediction, including whether and how motion contributes.
This need is not tied to a particular output format: whether a model predicts an action label or answers a language question, explanations should reveal the dynamic and static evidence behind the decision.

\begin{figure*}[t]
    \centering
    \includegraphics[width=\textwidth]{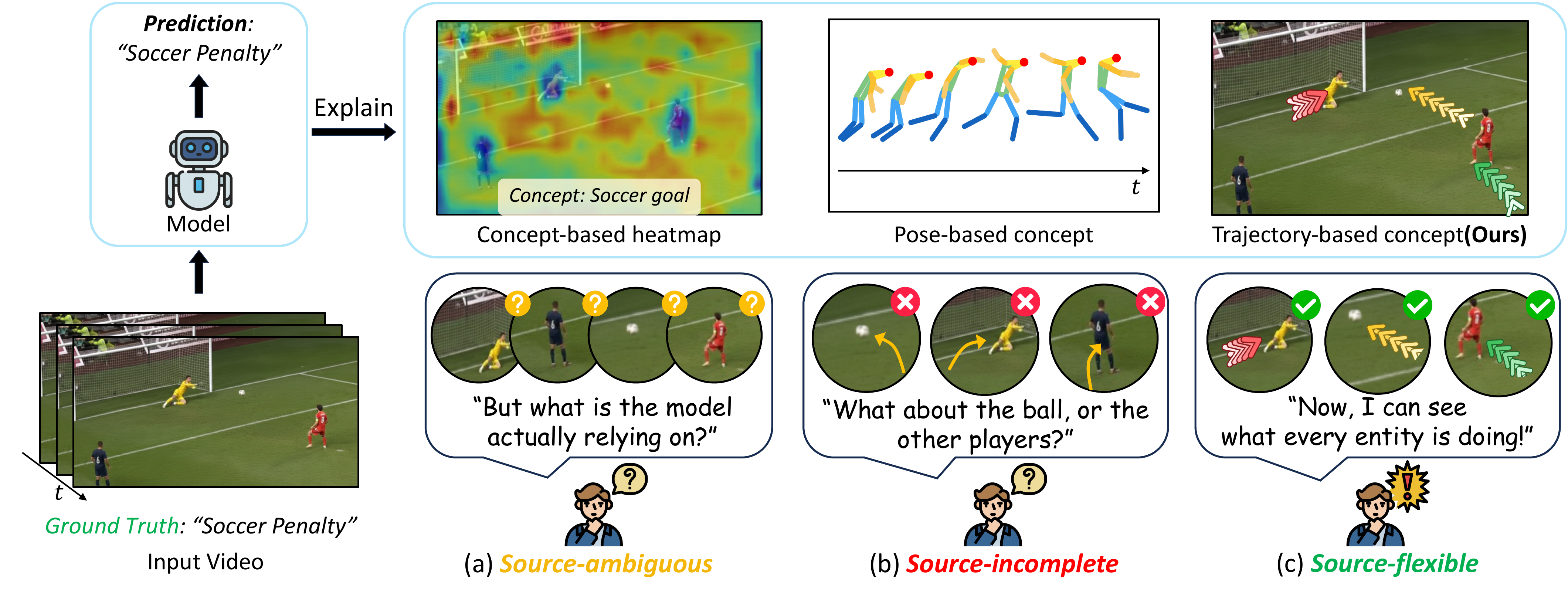        }
\vspace{\figcapmargin}
\caption{
\textbf{Trajectory concepts make motion sources flexible.}
Existing video XAI often under-specifies \emph{which motion source} supports a prediction.
(a) Heatmap-based explanations~\cite{salf-cbm} can conflate actor motion, object motion, and static context (\emph{source-ambiguous}).
(b) Pose-based concepts~\cite{dance} cover one person's joints but omit object motion and interactions (\emph{source-incomplete}).
(c) In contrast, \ours{} provides \emph{source-flexible motion explanations} by representing dynamic evidence from people, objects, and interactions as trajectory concepts.
}
\vspace{-2em}
    \label{fig:teaser}    
\end{figure*}

\vspace{-0.1em}
When motion contributes, explaining it as a single factor is insufficient.
Motion in videos is not monolithic: it can arise from actors, manipulated objects, or interactions among entities, each carrying different semantics~\cite{meunier2025segmenting,wang1994representing}.
For example, in a soccer play, a player's leg swing, the ball trajectory, and a defender's movement correspond to distinct sources of motion.
An explanation that fails to specify these sources suffers from \emph{motion-source under-specification}: it may indicate that ``motion'' matters, but not which motion supports the prediction.
This failure can take two forms, as shown in \figref{teaser}.
An explanation may conflate multiple motion sources into an ambiguous visual cue (\emph{source-ambiguous}), or omit important sources altogether (\emph{source-incomplete}).
In both cases, users cannot determine whether a model relies on action-relevant body dynamics, object trajectories, interactions, or static contextual shortcuts.
Thus, explainable video understanding requires motion explanations that specify dynamic evidence by source and present it distinctly.

\vspace{-0.1em}
Despite this need, existing video XAI methods suffer from \emph{motion-source under-specification}.
Attribution-based and region-centric methods~\cite{hartley2022swag,video-perturb,stergiou2019saliency} explain predictions by producing spatio-temporal heatmaps, tubelets, or other localized visual evidence over input videos.
While these outputs can indicate where influential evidence appears, they do not specify what type or source of evidence the model uses.
A highlighted region may correspond to static appearance, object or scene context, actor motion, object motion, or their mixture.
Thus, such explanations are \emph{source-ambiguous}: they do not explicitly represent motion as interpretable concepts, let alone separate motion concepts according to their sources.
For example, as shown in \figref{teaser} (a), highlighting a player and a ball cannot tell whether the model relies on the player's kicking motion, the ball trajectory, or static contextual cues.
This forces users to infer motion evidence from frame-to-frame changes in the visualization, making the explanation underspecified and ambiguous.

\vspace{-0.1em}
Concept-level explanations can reduce this ambiguity by exposing prediction evidence as interpretable concepts rather than raw spatio-temporal regions.
However, in video understanding, their reliability critically depends on how motion concepts are defined.
Language-derived concepts are effective for objects and scenes, but fine-grained motion is often \emph{tacit} and difficult to express in words~\cite{reber1989implicit}.
Recent pose-centric concept methods represent motion with human pose or skeleton sequences~\cite{dance,pcbear}, making single-person body motion explicit and interpretable.
This pose-centric formulation, however, is \emph{source-incomplete}: its motion vocabulary is defined over visible human joints and cannot directly represent dynamic evidence beyond single-person pose, such as object trajectories or multi-agent interactions.
As a result, motion from interacting agents or manipulated objects may be omitted, or only indirectly reflected through human-pose or static-context concepts.
This calls for motion concepts that are not tied to predefined human joints, but can instead trace dynamic evidence across diverse moving entities.

\vspace{-0.1em}
To address motion-source under-specification, we propose Trajectory-Aware Concepts for Explainable Video Understanding (\ours{}), a concept-based framework that treats visual trajectories as first-class explanatory concepts.
Derived from optical-flow trajectories, these concepts form a \emph{source-flexible vocabulary for dynamic evidence} from people, manipulated objects, and interactions.
As illustrated in \figref{teaser}, \ours{} explains predictions through these trajectory-aware motion concepts, while using object and scene concepts to account for static context.
Because these concepts are defined at the level of video evidence rather than task-specific labels, the same typed trajectory/object/scene interface can support both class prediction and question-conditioned answer scoring.
This shifts the explanatory bottleneck from describing single-person body poses to representing dynamic visual evidence at the trajectory level.
To our knowledge, \ours{} is the first concept-based video explanation framework that uses optical-flow trajectory concepts as a unified interface for explaining dynamic evidence across action recognition and VQA.

\vspace{-0.1em}
We instantiate \ours{} on two representative video understanding tasks: action recognition and video question answering (VQA).
Across six benchmarks---NTU RGB+D Mutual~\cite{ntu}, Chiral SSv2~\cite{ssv2,chirality}, UCF-101~\cite{ucf}, KTH~\cite{kth}, TOMATO~\cite{TOMATO}, and STAR~\cite{star}---we evaluate single-person body motion, multi-person interactions, object-centric dynamics, temporally directional cues, and language-grounded reasoning.
For VQA, we use native VQA benchmarks together with multiple-choice VQA reformulations of action-recognition datasets, allowing us to test whether the same typed interface supports both class prediction and question-conditioned answer scoring.
Spatial grounding analyses and user studies support sample-level
alignment with motion sources and human-perceived interpretability,
respectively.
\ours{} maintains competitive task performance, while its trajectory-only
variant retains predictive value under object-category shift and
shows modest cross-domain transfer.
Motion perturbation tests and concept removal provide complementary
evidence of directional semantic alignment and decision faithfulness,
while sample-level interventions illustrate targeted model debugging.

\vspace{-0.1em}
To summarize, we make the following three contributions.
\vspace{-0.1em}
\begin{itemize}

    \item We identify \emph{motion-source under-specification} as a
    limitation of existing video explanations and introduce
    \emph{trajectory concepts} as a source-flexible representation
    of dynamic evidence from people, objects, and interactions.

    \item We propose \ours{}, a concept-based framework that
    automatically discovers motion concepts by clustering optical-flow
    trajectories and complements them with object and scene concepts,
    without requiring manual concept annotation.
    The resulting typed interface supports sparse classification for
    action recognition and question-conditioned concept matching for VQA.

    \item Five-seed evaluations across six benchmarks show competitive
    task performance, with an average accuracy drop of 1.0 points for
    action recognition and a nominal gain of 0.4 points for VQA
    relative to non-interpretable baselines.
    Spatial grounding analyses and user studies support sample-level
    spatial alignment and interpretability, while perturbation and
    intervention tests provide complementary evidence of directional
    semantic alignment and decision faithfulness.
\end{itemize}
\vspace{\secmargin}

\section{Related Work}
\label{sec:related}

\vspace{\paramargin}
\paragraph{Video understanding and dynamic evidence.}
Modern video understanding models span spatio-temporal action recognizers~\cite{simonyan2014two,I3D,slowfast,timesformer,videomae,cast,ca2st} and video-language models for language-grounded reasoning~\cite{videollava,wang2024internvideo2,qwen3vl}.
Yet their predictions can depend on entangled dynamic motion and static context~\cite{modirect}.
Recent action-scene or spatial-temporal disentanglement methods address this issue for robustness or hallucination reduction~\cite{bae2024devias,mash-vlm,rcore}, but do not expose dynamic evidence as concepts or specify which motion source supports an individual prediction.
\ours{} instead explains dynamic evidence with trajectory concepts and static context with object and scene concepts.

\vspace{\paramargin}
\paragraph{Region-centric and post-hoc concept explanations in video XAI.}
A first form of motion-source under-specification arises in region-centric video XAI.
Saliency, perturbation, tube, and supervoxel methods localize where evidence appears~\cite{stergiou2019saliency,video-perturb,hartley2022swag}, but remain \emph{source-ambiguous}: a highlighted region may mix actor motion, object motion, static appearance, and scene context.
Post-hoc concept methods~\cite{3d-ace,vtcd,video-tcav} discover or probe recurring spatio-temporal concepts, but their concepts remain \emph{untyped} rather than reusable motion concepts.
\ours{} explains dynamic evidence at the trajectory level, avoiding source-ambiguous regions and generic spatio-temporal concepts.

\vspace{\paramargin}
\paragraph{Concept bottlenecks and pose-centric motion concepts.}
Concept bottleneck models explain predictions through human-interpretable concepts~\cite{cbm,lf-cbm,pcbm, labo,vlgcbm,ta-cbm,ECBM}, but video CBMs depend critically on how motion concepts are defined.
Language-derived concepts struggle with fine-grained motion, while pose-centric methods~\cite{dance,pcbear} restrict motion concepts to human pose or skeleton sequences.
Such vocabularies are \emph{source-incomplete}: they can explain single-person body motion but cannot directly represent object trajectories or multi-agent interactions.
\ours{} addresses this limitation with optical-flow trajectory concepts, providing a source-flexible motion vocabulary.

\vspace{\paramargin}
\paragraph{Explainable video-language understanding.}
Beyond action recognition, video-language works improve interpretability through temporal or spatial grounding~\cite{nextgqa,timecraft,igv,deng2025motion,evident} and textual or structural reasoning traces~\cite{tvtrees,vget,chen2021explainable,chen2024mecd}.
These signals localize evidence or organize reasoning, but do not provide concept-level explanations of what dynamic evidence the model uses.
In particular, they do not expose reusable motion concepts or distinguish the motion source behind a prediction.
\ours{} represents dynamic evidence with trajectory concepts, yielding a task-general explanation interface for action recognition and VQA.

\section{\ours{}: A Trajectory-Aware Concept Interface}
\label{sec:method}
\begin{figure*}[t]
\includegraphics[width=\textwidth]{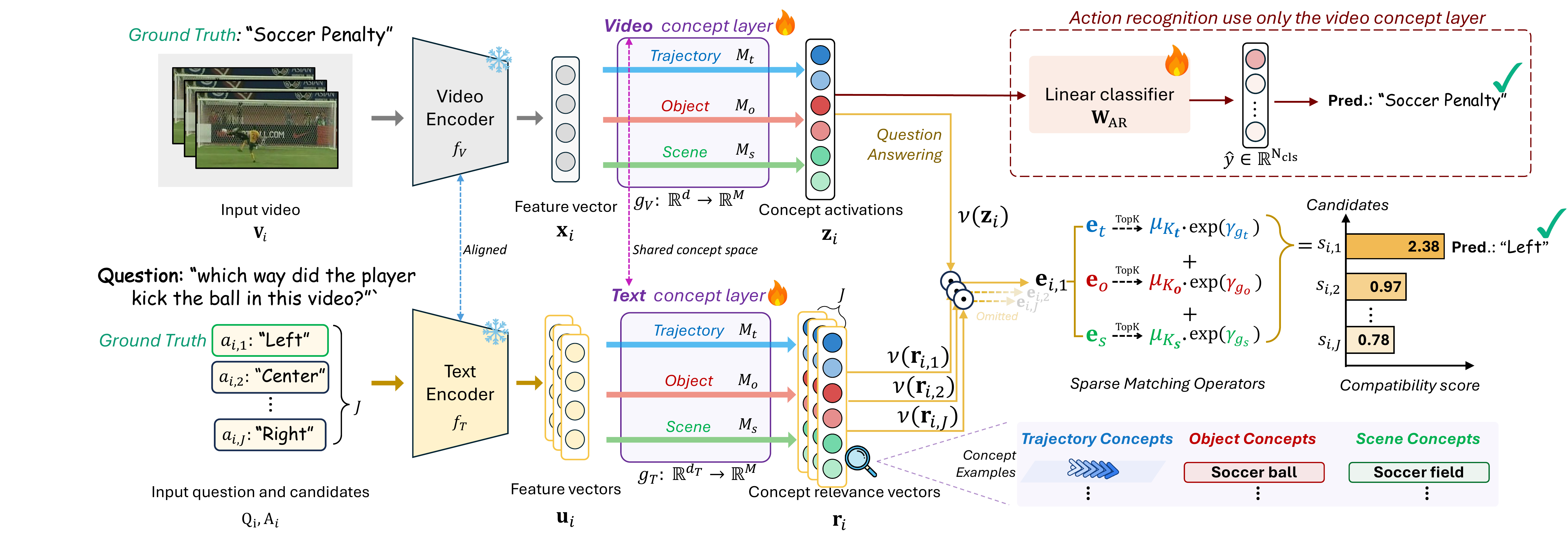}
\caption{
\textbf{Overview of \ours{}.}
\ours{} represents videos with a typed trajectory/object/scene concept interface: trajectory concepts capture dynamic evidence, while object and scene concepts capture static context.
Action recognition predicts directly from video-side concept activations, whereas VQA projects each question--answer pair into concept relevance and matches it with the video-side concepts.
Thus, the same typed interface supports trajectory-based explanations for both class prediction and language-grounded reasoning.
}
\label{fig:overview}
\vspace{\figmargin}
\end{figure*}

\subsection{Design Overview}
\label{sec:design_overview}

\vspace{\paramargin}
\ours{} explains video understanding models through the trajectory-aware concept interface in \figref{overview}.
The interface separates dynamic evidence from static context and exposes predictions through three typed concept groups:
for a video \(\mathbf{V}_i\),
$
    \mathbf{z}_i =
    [\mathbf{z}^{t}_i;\mathbf{z}^{o}_i;\mathbf{z}^{s}_i],
$
where \(\mathbf{z}^{t}_i\) denotes trajectory concept activations, and \(\mathbf{z}^{o}_i,\mathbf{z}^{s}_i\) denote object and scene concept activations.
The design follows three principles.

\vspace{\paramargin}
\paragraph{D1. Visual trajectories as motion concepts.}
\ours{} treats \emph{visual trajectories as first-class explanatory concepts}.
Unlike pose-centric motion concepts tied to predefined human joints, trajectory concepts are derived from optical-flow trajectories and capture visible entity-level dynamics from actors, manipulated objects, and interactions among multiple entities.
This yields a source-flexible motion vocabulary without manual motion annotation or natural-language motion descriptions.

\vspace{\paramargin}
\paragraph{D2. A typed interface for dynamic and static evidence.}
As illustrated in \figref{overview}, trajectory concepts explain dynamic evidence, while object and scene concepts account for static context.
The exact concept vocabularies are discovered separately from each downstream dataset's training videos, but the typed interface---trajectory, object, and scene---remains fixed.
This allows \ours{} to adapt its concept vocabulary to each dataset while preserving a consistent explanation structure.

\vspace{\paramargin}
\paragraph{D3. Task-general consumption of the concept interface.}
The typed video-side concept interface supports different video understanding tasks.
For action recognition, a sparse classifier predicts the action label directly from \(\mathbf{z}_i\).
For VQA, each question--answer pair is projected into the same concept space as a question-conditioned concept relevance vector and matched with \(\mathbf{z}_i\).
Thus, VQA prediction becomes concept-level: the model answers by matching concepts present in the video with concepts required by the candidate answer.
Importantly, \emph{trajectory concepts do not require language names}.
This is crucial for VQA: although trajectory concepts are visual motion clusters, answer supervision can still teach the text branch which concepts are needed for a candidate answer.
Object and scene relevance, in contrast, can be supervised using language concept names.

\vspace{-0.3em}
In the following subsections, we present task-agnostic concept discovery (\secref{concept_discovery}), concept bottleneck learning (\secref{concept_learning}), and task instantiations (\secref{task_instantiation}).
\vspace{\secmargin}

\subsection{Task-Agnostic Concept Vocabulary Discovery}
\label{sec:concept_discovery}
\begin{figure*}[t]
\centering
\includegraphics[width=\textwidth]{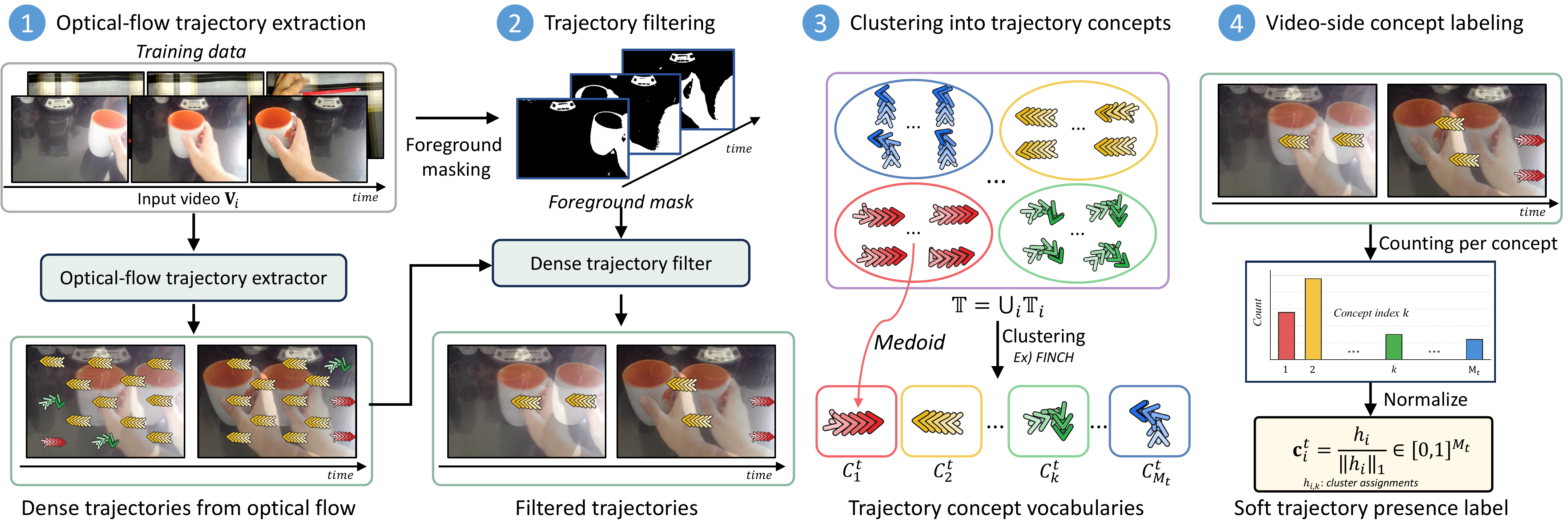}
\vspace{\figcapmargin}
\figcaption{Trajectory concept discovery and labeling}
{
\ours{} extracts optical-flow trajectories from training videos, filters weak or noisy motion, and clusters the remaining trajectories into recurring motion patterns.
Each cluster defines a trajectory concept, and each video is labeled by the normalized frequency of its trajectory assignments.
This yields a \emph{source-flexible motion vocabulary} for people, objects, and interactions without pose annotations or language descriptions of motion.
}
\label{fig:concept_discovery}
\vspace{\figmargin}
\end{figure*}

\vspace{\paramargin}
For each downstream dataset, \ours{} discovers concept vocabularies from its training videos, independently of whether the final task is action recognition or VQA.
These vocabularies consist of trajectory concepts \(\mathbb{C}^{t}\) for entity-level dynamic evidence, and object and scene concepts \(\mathbb{C}^{o},\mathbb{C}^{s}\) for static context.
All concepts are discovered automatically without manual concept annotation.
\vspace{\secmargin}

\subsubsection{Trajectory Concepts}
\label{sec:trajectory_concepts}

\paragraph{Trajectory generation.} 

Given a video \(\mathbf{V}_i\), we estimate dense optical flow between consecutive frames using a pretrained flow estimator~\cite{waft}.
We divide the video into \(N_s\) temporal segments and extract an \(L\)-frame clip around each segment center.
For one segment, let $\Delta\mathbf{p}_r$ denote the optical flow from frame \(r\) to \(r+1\).
Starting from a seed pixel \(\mathbf{p}_0\in\mathbb{R}^2\), we trace a visual trajectory by accumulating flow as illustrated in \figref{concept_discovery},
\begin{equation}
    \Delta\mathbf{p}_r = \mathbf{p}_{r+1} - \mathbf{p}_{r},
    \qquad
    \tau(\mathbf{p}_0)   = [\Delta\mathbf{p}_0,\ldots,\Delta\mathbf{p}_{L-2}] \in \mathbb{R}^{2(L-1)}.
    \label{eq:flow}
\end{equation}
Applying this to all segments yields raw trajectories.
We retain foreground trajectories with large motion magnitude, producing a filtered trajectory set \(\mathbb{T}_i\) for video \(\mathbf{V}_i\).
Details of interpolation, foreground masking, and trajectory filtering are provided in \apref{trajectory_concept}.

\vspace{\paramargin}
\paragraph{Trajectory concept discovery and labeling.}
We aggregate trajectories over the training set, \(\mathbb{T}=\bigcup_i \mathbb{T}_i\), and cluster them with a clustering algorithm.
\ours{} is clustering-agnostic; we use FINCH~\cite{finch} in our experiments.
Let \(\pi:\mathbb{T}\rightarrow\{1,\ldots,M_t\}\) denote the cluster assignment.
The \(k\)-th trajectory concept is
\begin{equation}
    C_k^t=\{\tau\in \mathbb{T}\mid \pi(\tau)=k\},
    \qquad
    \mathbb{C}^t=\{C_k^t\}_{k=1}^{M_t}.
    \label{eq:traj_concept}
\end{equation}
Each concept captures a recurring trajectory-level motion pattern shared across videos.
For video \(\mathbf{V}_i\), we count cluster assignments
\(h_{i,k}=\sum_{\tau\in \mathbb{T}_i}\mathbbm{1}[\pi(\tau)=k]\)
and form a soft trajectory presence label $\mathbf{c}^t_i=\frac{\mathbf{h}_i}{\|\mathbf{h}_i\|_1} \in [0,1]^{M_t}.$
Thus, \(c^t_{i,k}\) indicates the relative frequency of trajectory concept \(k\) in \(\mathbf{V}_i\).
\vspace{\secmargin}

\subsubsection{Object and Scene Concepts}
\label{sec:object_scene_concepts}

\vspace{\paramargin}
To account for static context, we use object and scene concepts.
Unlike prior class-name prompting~\cite{dance}, we query training videos with a Video-LLM~\cite{qwen3vl}, so the vocabularies reflect visible evidence rather than action-category priors.
After filtering long or redundant phrases, we obtain \(\mathbb{C}^{o}\) and \(\mathbb{C}^{s}\).
We assign soft pseudo-labels by video-text similarity using a pretrained vision-language encoder~\cite{wang2024internvideo2}.
Additional details on prompts, filtering, and score normalization are provided in \apref{object_and_scene_concept}.
\vspace{\secmargin}

\subsection{Concept Bottleneck Learning}
\label{sec:concept_learning}

\vspace{\paramargin}
Given a video \(\mathbf{V}_i\), a frozen video encoder \(f_V\) extracts
\(\mathbf{x}_i=f_V(\mathbf{V}_i)\in\mathbb{R}^{d}\).
We define a video-side concept layer \(g_V:\mathbb{R}^{d}\rightarrow\mathbb{R}^{M}\), where \(M=M_t+M_o+M_s\) is the total number of trajectory, object, and scene concepts.
Then we compute raw concept scores $\mathbf{a}_i=g_V(\mathbf{x}_i)  = [\mathbf{a}^{t}_i;\mathbf{a}^{o}_i;\mathbf{a}^{s}_i]$ and activations $\mathbf{z}_i= [\sigma(\mathbf{a}^{t}_i);\mathbf{a}^{o}_i;\mathbf{a}^{s}_i]$.
The sigmoid is applied only to trajectory concepts, which are trained with soft-label binary cross-entropy \(L_t\) using the soft trajectory label \(\mathbf{c}^t_i\).
Object and scene branches are trained with cosine-cubed losses \(L_o,L_s\) using pseudo-labels \(\mathbf{c}^o_i,\mathbf{c}^s_i\), following label-free concept bottleneck models~\cite{lf-cbm,dance}.
The video-side concept learning objective is
\begin{equation}
    L_{\mathrm{concept}}
    =
    \lambda_t L_t+\lambda_o L_o+\lambda_s L_s.
    \label{eq:loss_concept}
\end{equation}
This stage initializes the video-side concept interface.
For action recognition, we freeze the learned video-side concept layer and train only the sparse classifier.
For VQA, we freeze the video encoder but continue updating the video-side concept layer under concept supervision, while learning the text-side concept projector and answer-scoring parameters.
\vspace{\secmargin}

\subsection{Task Instantiations}
\label{sec:task_instantiation}

\vspace{\paramargin}
For each downstream dataset, \ours{} discovers its own trajectory, object, and scene concept vocabularies from training videos.
Thus, the exact concepts are task-specific, while the typed interface $\mathbf{z}_i=[\mathbf{z}^{t}_i;\mathbf{z}^{o}_i;\mathbf{z}^{s}_i]$ is shared, as shown in \figref{overview}.
Action recognition consumes this interface directly with a sparse classifier, while VQA matches it with question-conditioned concept relevance.

\vspace{\secmargin}
\subsubsection{Action Recognition}
\label{sec:ar}

\vspace{\paramargin}
For action recognition, each sample is \((\mathbf{V}_i,y_i)\), where \(y_i\in\{1,\ldots,N_{\mathrm{cls}}\}\) is the action label.
We freeze the video-side concept layer and train a sparse linear classifier on \(\mathbf{z}_i\):
\(\hat{\mathbf{y}}_i=\mathrm{softmax}(\mathbf{W}_{\mathrm{AR}}\mathbf{z}_i)\).
We optimize cross-entropy with elastic-net regularization~\cite{sparse} on \(\mathbf{W}_{\mathrm{AR}}\), encouraging each prediction to rely on a compact set of trajectory, object, and scene concepts.

\vspace{\secmargin}
\subsubsection{Video Question Answering via Concept Matching}
\label{sec:vqa}

For VQA, each sample is \((\mathbf{V}_i,Q_i,A_i,y_i)\), where \(Q_i\) is a question,
\(A_i=\{a_{i1},\ldots,a_{iJ}\}\) is the set of answer candidates, and \(y_i\) is the index of the correct answer.

Our VQA instantiation builds on the concept-space matching design of prior work~\cite{mm-cbm}, but adapts it to video understanding through \emph{a typed trajectory/object/scene interface} and \emph{a name-free trajectory branch}.

\vspace{\paramargin}
\paragraph{Text-side trajectory relevance without motion names.}
The key challenge is that trajectory concepts are visual motion clusters \emph{without natural-language names}.
Object and scene relevance can be supervised by concept-name similarity, but trajectory relevance cannot.
We therefore learn text-side trajectory relevance through answer-supervised alignment with video-side trajectory evidence, rather than through text-concept similarity.

\vspace{\paramargin}
\paragraph{Question--answer projection.}
Let \(f_T\) be a pretrained text encoder and \(g_T:\mathbb{R}^{d_T}\rightarrow\mathbb{R}^{M}\) a text-side concept projector.
For each candidate answer, we compute question-conditioned concept relevance:
\begin{equation}
    \mathbf{r}_{ij}
    =
    g_T(f_T([Q_i;a_{ij}]))
    =
    [\mathbf{r}^{t}_{ij};\mathbf{r}^{o}_{ij};\mathbf{r}^{s}_{ij}].
    \label{eq:concept_rel}
\end{equation}    
Here, \(\mathbf{z}_i\) represents video-side concepts, while \(\mathbf{r}_{ij}\) represents concepts needed to support answer \(a_{ij}\).

\vspace{\paramargin}
\paragraph{Concept-level answer scoring.}
We measure a concept-compatibility score of each answer as:
\begin{equation}
    s_{ij}
    =
    \sum_{g\in\{t,o,s\}}
    \exp(\gamma_g)
    \mu_{K_g}(\mathbf{z}^{g}_i,\mathbf{r}^{g}_{ij}),
    \label{eq:qa_score}
\end{equation}
where \(\mu_{K_g}\) sums the top-\(K_g\) entries of \(\nu(\mathbf{z})\odot\nu(\mathbf{r})\) (\(\odot\) denotes element-wise multiplication), following prior works~\cite{mm-cbm, vlgcbm}, \(\nu(\mathbf{x})=\mathrm{ReLU}(\mathbf{x})/(\|\mathrm{ReLU}(\mathbf{x})\|_2+\epsilon)\), and \(\gamma_g\) is a learnable weight of the concept group $g$.
The predicted answer is \(a_{ij_i^{\mathrm{pred}}}\), where \(j_i^{\mathrm{pred}}=\arg\max_j s_{ij}\).

\vspace{\paramargin}
\paragraph{VQA training objective.}
To preserve concept fidelity, we retain the video-side concept loss $L_{\mathrm{concept}}$, computed over videos in each mini-batch.
The answer loss is the cross-entropy over candidate scores, denoted \(L_{\mathrm{QA}}\).
For object and scene concepts, we supervise text-side relevance with a candidate-averaged cosine loss \(L_{\mathrm{text}}^g\), \(g\in\{o,s\}\), using text-concept similarity labels.
For trajectory concepts, text-concept similarity is unavailable, so we align question-conditioned trajectory relevance with video-side trajectory evidence using an InfoNCE loss~\cite{infonce} \(L_{\mathrm{align}}^t\), where the correct answer is the positive and other candidates are negatives.
The overall VQA objective is
\begin{equation}
    L_{\mathrm{VQA}}
    =
    \lambda_{\mathrm{QA}}L_{\mathrm{QA}}
    +
    \lambda_{\mathrm{concept}}L_{\mathrm{concept}}
    +
    \lambda_{\mathrm{text}}(L_{\mathrm{text}}^o + L_{\mathrm{text}}^s)
    +
    \lambda_{\mathrm{align}} L_{\mathrm{align}}^t.
    \label{eq:loss_vqa}
\end{equation}
In summary, $L_{\mathrm{concept}}$ preserves the video-side concept semantics during VQA training, 
$L_{\mathrm{text}}^g$ trains text-side relevance for \emph{named} object and scene concepts, 
and $L_{\mathrm{align}}^t$ connects \emph{unnamed} trajectory concepts to question-conditioned answer relevance.

\vspace{\secmargin}

\section{Experimental Results}
\label{sec:results}

\vspace{\paramargin}

We assess explanation interpretability and sample-level grounding (\secref{analysis}), task performance and generalization (\secref{performance}), and concept semantics and decision faithfulness (\secref{edit}).
Dataset and implementation details are provided in \apref{dataset} and \apref{impl}, respectively.

\vspace{\secmargin}

\subsection{Source-Flexible Explanations with Trajectory Concepts}
\label{sec:analysis}

\vspace{\paramargin}
\paragraph{Concept contribution and visualization.}
For action recognition, the contribution of concept \(m\) to the predicted class \(\hat{c}_i=\arg\max_c \hat{\mathbf{y}}_i[c]\) is
\(z_{i,m}\mathbf{W}_{\mathrm{AR}}[\hat{c}_i,m]\), the product of its activation and classifier weight.
For VQA, we use the compatibility score \(s_{ij}\): each concept contributes through top-\(K\) matching between video-side evidence \(z_i^g[m]\) and question-conditioned relevance \(r_{ij}^g[m]\).
We visualize trajectory concepts using representative cluster trajectories overlaid on top-activated clips with Grad-CAM~\cite{gradcam}, used \emph{only to localize responses for visualization}.

\paragraph{\ours{} provides source-flexible motion explanations.}

\begin{wraptable}{r}{0.28\linewidth}
\vspace{-4mm}
\centering
\caption{\textbf{Motion-source grounding.} Top-10 concept support within annotated boxes (\%).}
\label{tab:source_grounding_main}
\resizebox{0.95\linewidth}{!}{%
\begin{tabular}{lcc}
\toprule
Method & Hand & Object \\
\midrule
DANCE~\cite{dance} & 42.3 & 7.6 \\
\ours{}             & \textbf{64.5} & \textbf{63.5} \\
\bottomrule
\end{tabular}%
}
\vspace{-1.5mm}
\end{wraptable}

We quantify sample-level spatial grounding on object-disjoint Chiral SSv2 using object bounding box annotations from Something-Else dataset~\cite{somethingelse}.
For this post-hoc analysis, we associate the top-contributing concepts with input trajectories through cluster assignment.
As shown in \tabref{source_grounding_main}, \ours{} achieves superior coverage of annotated hand and manipulated-object boxes than DANCE.
These results support spatial alignment with both entity types without fixed entity labels for trajectory concepts.
Protocol details and additional concept-selection results are provided in \apref{source_grounding}.

\begin{figure}[t]
    \centering
    \includegraphics[width=\textwidth]{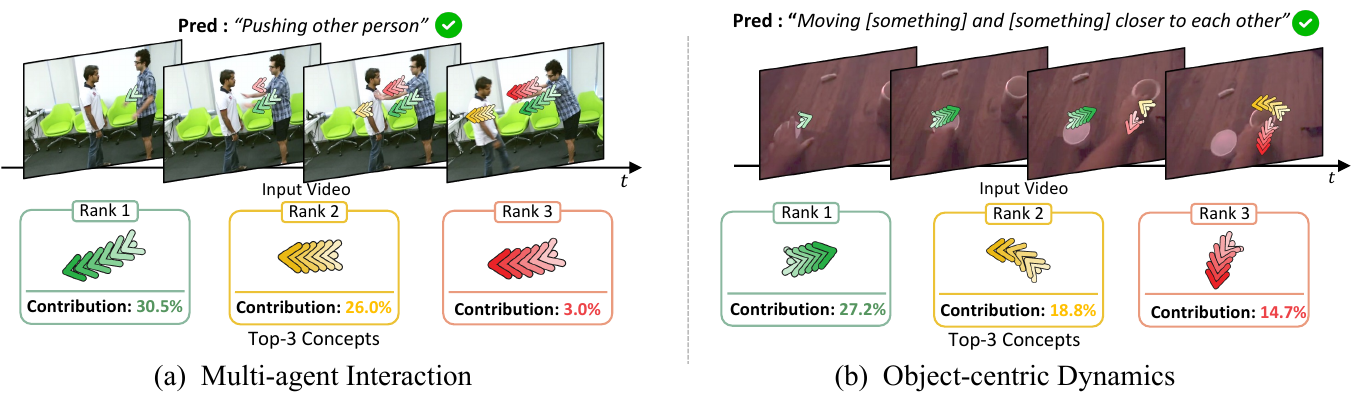}
    \vspace{\figcapmargin}
    \caption{\textbf{Sample-level explanation.} 
    We visualize the top-3 contributing concepts by overlaying them on top-activated video regions.
    (a) Interaction dynamics across agents.
    (b) Object-centric motion with partial human cues.
    Trajectory concepts capture motion from multiple sources.
    }
    \label{fig:contrib}
    \vspace{\figmargin}
\end{figure}

\noindent
Complementary visualizations in \figref{contrib} show the top-3 contributing trajectory concepts for motion beyond single-person pose dynamics.
For \textit{pushing other person}, the selected concepts correspond to both the pushing motion of one person and the displacement of the other.
For \textit{moving [something] and [something] closer to each other}, they correspond to the objects' converging movements even though no full human body is visible.

\vspace{\paramargin}

\begin{figure*}[t]
    \centering
    \includegraphics[width=\textwidth]{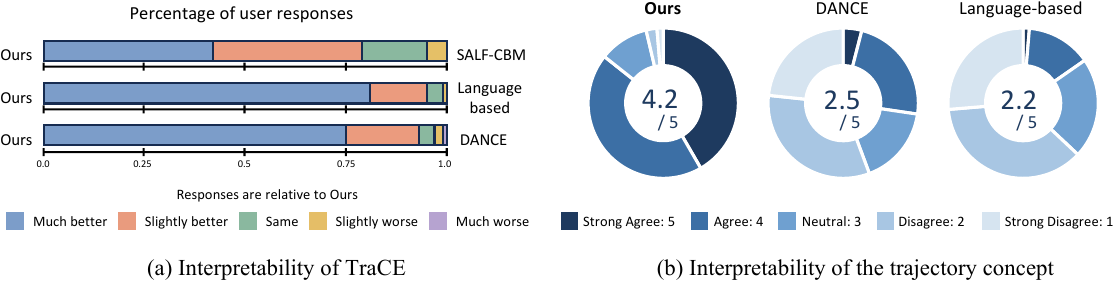}
    \vspace{\figcapmargin}
    \caption{
    \textbf{Interpretability of \ours{}.}
    (a) Pairwise preference study comparing \ours{} explanations with SALF-CBM~\cite{salf-cbm}, a language-based motion-concept CBM, and DANCE~\cite{dance}. 
    Bars show responses relative to \ours{}.
    (b) Concept-level study comparing trajectory concepts (\ours{}), pose-sequence concepts (DANCE), and language-derived motion concepts.
    Participants rate how well each concept matches the motion shared by its top-activating video clips on a five-point Likert scale.
    }
    \label{fig:us1}
    \vspace{\figmargin}
\end{figure*}

\paragraph{Trajectory concepts and explanations are interpretable.}
We conduct two user studies on motion-focused subsets of NTU RGB+D~\cite{ntu} and Chiral SSv2~\cite{ssv2,chirality} (\apref{userstudy_details}).
First, participants compare \ours{} explanations against DANCE~\cite{dance}, a language-based motion-concept CBM, and SALF-CBM~\cite{salf-cbm}.
As shown in \figref{us1} (a), at least 93\% of responses prefer \ours{} over DANCE and the language-based baseline, and about 78\% prefer \ours{} over SALF-CBM.
Second, participants rate how well a concept matches the motion shared by its top-4 activating clips.
As shown in \figref{us1} (b), trajectory concepts achieve the highest score (4.2/5), compared to DANCE pose concepts (2.5/5) and language-based concepts (2.2/5).
These results indicate that users find trajectory concepts coherent and better aligned with the motion patterns in the clips.

\begin{figure}[t]
    \centering
    \begin{minipage}[t]{0.48\linewidth}
        \centering
        \includegraphics[width=\linewidth]{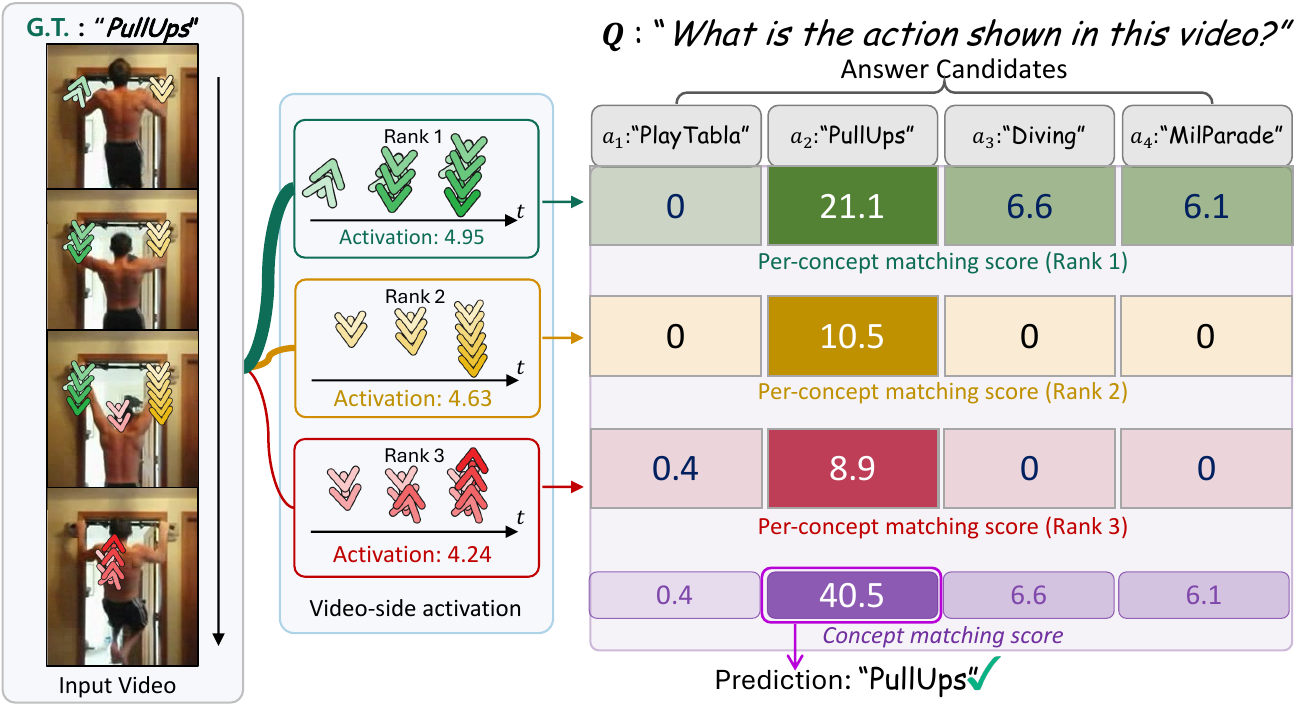}
        \vspace{\figcapmargin}
        \captionof{figure}{
        \textbf{Sample-level VQA explanation.}
        Rows show the top-3 video-side concepts and columns show answer candidates, with cells indicating concept--answer compatibility.
        }
        \label{fig:qa}
    \end{minipage}
    \hfill
    \begin{minipage}[t]{0.50\linewidth}
        \centering
        \includegraphics[width=\linewidth]{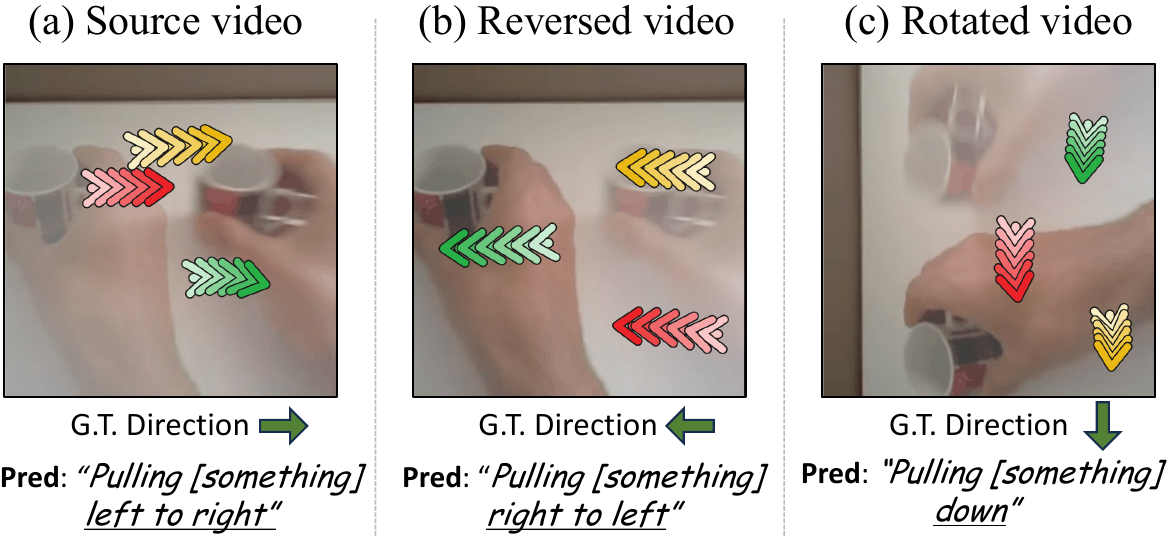}
        \vspace{\figcapmargin}
        \captionof{figure}{
       \textbf{Motion perturbation analysis.}
    Activated trajectory concepts track directional changes under reversal and rotation.
    Each panel overlays the first frame at low opacity with the last frame at full opacity.
}
        \label{fig:sanity}
    \end{minipage}
    \vspace{\figmargin}

\end{figure}

\vspace{\paramargin}

\paragraph{Concept matching explains language-grounded predictions.}
\figref{qa} shows how \ours{} explains a VQA prediction by matching video-side concepts with question-conditioned concept relevance.
The correct answer \textit{PullUps} receives high scores across the top-3 trajectory concepts, which capture repeated vertical motion, while competing answers receive lower scores.
To assess whether these concept-based explanations help users understand the model's answers, we conduct a blinded pairwise study with 50 participants.
We control for answer correctness by including only examples for which both compared models predict the same correct answer.
Participants prefer \ours{} explanations over those based on language-derived motion concepts and pose-sequence concepts in 70.4\% and 85.6\% of comparisons, respectively, indicating greater perceived usefulness in the VQA setting.
The full protocol and response distributions are provided in \apref{vqa_user_study}.

\vspace{\paramargin}

\vspace{\secmargin}

\subsection{Performance and Generalization of the Trajectory-Aware Interface}
\label{sec:performance}
\vspace{\paramargin}

We next assess the predictive performance of the typed trajectory/object/scene interface across action recognition and VQA, including transfer under object-category and domain shifts.

\begin{table*}[t]
\centering
\caption{
\textbf{Performance comparison on action recognition and VQA benchmarks.}
We report mean $\pm$ sample standard deviation over five random seeds for Top-1 accuracy (\%) using the same frozen backbone within each task, except for the single-run SpLiCE adaptation.
A dash (--) indicates that the method is not applicable to the task.
Task averages are computed over the corresponding dataset means.
}
\label{tab:main_results}
\resizebox{0.98\linewidth}{!}{
\begin{tabular}{l cc cc cc c cc cc}
\toprule
\multirow{2}{*}{Method}
& \multicolumn{2}{c}{NTU RGB+D Mutual}
& \multicolumn{2}{c}{Chiral SSv2}
& \multicolumn{2}{c}{UCF-101}
& KTH
& TOMATO
& STAR
& \multicolumn{2}{c}{\cellcolor{gray!20}Avg.} \\
\cmidrule(lr){2-3}
\cmidrule(lr){4-5}
\cmidrule(lr){6-7}
\cmidrule(lr){8-8}
\cmidrule(lr){9-9}
\cmidrule(lr){10-10}
\cmidrule(lr){11-12}
& AR & VQA
& AR & VQA
& AR & VQA
& AR
& VQA
& VQA
& \cellcolor{gray!20}AR
& \cellcolor{gray!20}VQA \\
\midrule
Baseline w/o interpretability
& $99.7{\pm}0.2$ & $97.4{\pm}0.1$
& $86.6{\pm}0.2$ & $82.6{\pm}0.2$
& $89.8{\pm}0.1$ & $99.8{\pm}0.0$
& $91.1{\pm}0.3$
& $35.9{\pm}2.2$
& $54.5{\pm}0.4$
& \cellcolor{gray!20}91.8
& \cellcolor{gray!20}74.0 \\
\midrule
LF-CBM~\cite{lf-cbm} w/ language concepts
& $96.9{\pm}0.3$ & --
& $77.6{\pm}0.5$ & --
& $88.3{\pm}0.2$ & --
& $89.1{\pm}0.7$
& --
& --
& \cellcolor{gray!20}88.0
& \cellcolor{gray!20}-- \\

DANCE~\cite{dance}
& $97.1{\pm}0.3$ & --
& $81.7{\pm}0.3$ & --
& $89.0{\pm}0.2$ & --
& $91.3{\pm}0.4$
& --
& --
& \cellcolor{gray!20}89.8
& \cellcolor{gray!20}-- \\

SALF-CBM~\cite{salf-cbm}
& $99.7{\pm}0.0$ & --
& $80.4{\pm}0.4$ & --
& $87.5{\pm}0.2$ & --
& $87.9{\pm}0.4$
& --
& --
& \cellcolor{gray!20}88.9
& \cellcolor{gray!20}-- \\

SpLiCE~\cite{splice} adaptation
& -- & 53.0
& -- & 57.3
& -- & 86.7
& --
& 22.4
& 49.2
& \cellcolor{gray!20}--
& \cellcolor{gray!20}53.7 \\

\ours{}
& $97.4{\pm}0.1$ & $97.1{\pm}0.3$
& $84.9{\pm}0.4$ & $82.6{\pm}0.6$
& $89.0{\pm}0.3$ & $99.7{\pm}0.1$
& $91.8{\pm}0.8$
& $37.0{\pm}1.7$
& $55.4{\pm}1.2$
& \cellcolor{gray!20}90.8
& \cellcolor{gray!20}74.4 \\
\bottomrule
\end{tabular}
}
\vspace{\tabmargin}
\end{table*}

\vspace{\paramargin}
\paragraph{\ours{} preserves performance across video understanding tasks.}
In \tabref{main_results}, we compare \ours{} with non-interpretable baselines using the same frozen backbone within each task: VideoMAE~\cite{videomae} for action recognition and InternVideo2~\cite{wang2024internvideo2} for VQA.
Across five random seeds, \ours{} remains close to the baseline on average, with a 1.0-point drop for action recognition and a nominal 0.4-point gain for VQA; given the reported variability, we do not interpret the VQA difference as a significant improvement.
We also adapt SpLiCE~\cite{splice} to question-conditioned answer scoring under the same frozen backbones.
This single-run adaptation performs below the five-seed mean of \ours{} on all five VQA benchmarks.
Overall, the concept interface retains competitive task performance while providing concept-based explanations.

\vspace{\paramargin}
\paragraph{Trajectory concepts remain predictive across diverse motion sources.}
We compare \ours{} with LF-CBM~\cite{lf-cbm} using language-derived motion concepts, DANCE using pose-sequence concepts~\cite{dance}, and SALF-CBM~\cite{salf-cbm} on action recognition.
Their original action-recognition formulations do not define question-conditioned answer scoring for VQA.
Across benchmarks spanning single-person motion (KTH), multi-person interaction (NTU RGB+D Mutual), object-centric and direction-sensitive dynamics (Chiral SSv2), and general human actions (UCF-101), \ours{} remains competitive among interpretable methods.
It achieves the highest reported mean accuracy on KTH and Chiral SSv2, ties for the best on UCF-101, and remains competitive on the near-saturated NTU RGB+D Mutual split.

\vspace{\paramargin}

\begin{wraptable}{r}{0.4\linewidth}
\vspace{-3mm}
\centering
\caption{
    \textbf{Generalization under object-category and visual-domain shifts.}
}
\label{tab:generalizability}
\resizebox{\linewidth}{!}{%
    \begin{tabular}{lcc}
    \toprule
    \multicolumn{1}{l}{\multirow{2}{*}{Method}} & Sth.-Else     & SSv2$\rightarrow$  \\
                                                & disjoint-obj. & TOMATO \\
    \midrule
    Baseline w/o interpretability            & 89.5 & 41.5  \\
    \ours{} w/o static concepts              & 94.4 & 43.1 \\
    \bottomrule
    \end{tabular}
}
\end{wraptable}

\paragraph{Trajectory concepts remain predictive under object-category and visual-domain shifts.}
We further evaluate whether the trajectory-only variant remains predictive under object-category and visual-domain shifts.
Using Something-Else object annotations~\cite{somethingelse}, we construct an object-disjoint Chiral SSv2~\cite{ssv2,chirality} split where object categories do not overlap between training and test.
As shown in \tabref{generalizability}, the trajectory-only variant of \ours{} achieves 94.4\%, outperforming the non-interpretable baseline~\cite{videomae} at 89.5\%.
We also train the concept pipeline on Chiral SSv2 and apply it to aligned TOMATO~\cite{TOMATO} examples without target-domain training.
The trajectory-only variant achieves 43.1\%, compared to 41.5\% for the baseline.
Together, these results suggest that trajectory concepts retain predictive motion cues under object category shift and show modest cross-domain transfer.

\vspace{\secmargin}
\subsection{Faithfulness of \ours{}}
\label{sec:edit}

\vspace{\paramargin}
We examine the directional semantics of activated trajectory concepts. 
Interpretability alone does not establish that concept explanations reflect model behavior.
We therefore examine whether activated concepts retain their directional meaning and whether the model's predictions depend on them, before illustrating their use for debugging.

\begin{wraptable}{r}{0.3\linewidth}
\vspace{-4mm}
\centering
\caption{\textbf{Prediction consistency under perturbations.}}
\label{tab:prediction_consistency_avg}

\resizebox{0.6\linewidth}{!}{%
\begin{tabular}{lc}
\toprule
Perturbation & Acc. \\
\midrule
Original & 81.1 \\
\midrule
Reverse  & 79.2 \\
Rotation & 63.1 \\
\bottomrule
\end{tabular}
}
\vspace{-1.5mm}
\end{wraptable}

\vspace{\paramargin}
\paragraph{Trajectory concepts respond consistently to motion perturbations.}
We examine sensitivity to signed motion direction at both the prediction and concept levels on the directional-class subset of Chiral SSv2~\cite{ssv2,chirality}, whose labels specify motion directions.
Temporal reversal swaps opposite directions, while a $90^\circ$ rotation maps each direction to its rotated counterpart.
As shown in \tabref{prediction_consistency_avg}, \ours{} achieves 81.1\% accuracy on original inputs, 79.2\% consistency under temporal reversal, and 63.1\% under rotation.
To assess concept semantics directly, we assign each concept a fixed direction label from the net displacement of its training-cluster medoid.
The highest-activated concept matches the ground-truth direction in 67.5\% of held-out videos, compared with a 25\% uniform four-way reference.
Qualitative examples in \figref{sanity} show that activated trajectory concepts track directional changes under temporal reversal and rotation.
Top-5 concept replacement and direction agreement are reported in \apref{directional_semantics}.
These analyses support motion sensitivity and directional semantic alignment; decision dependence is tested below.

\begin{figure*}[t]
    \includegraphics[width=\textwidth]{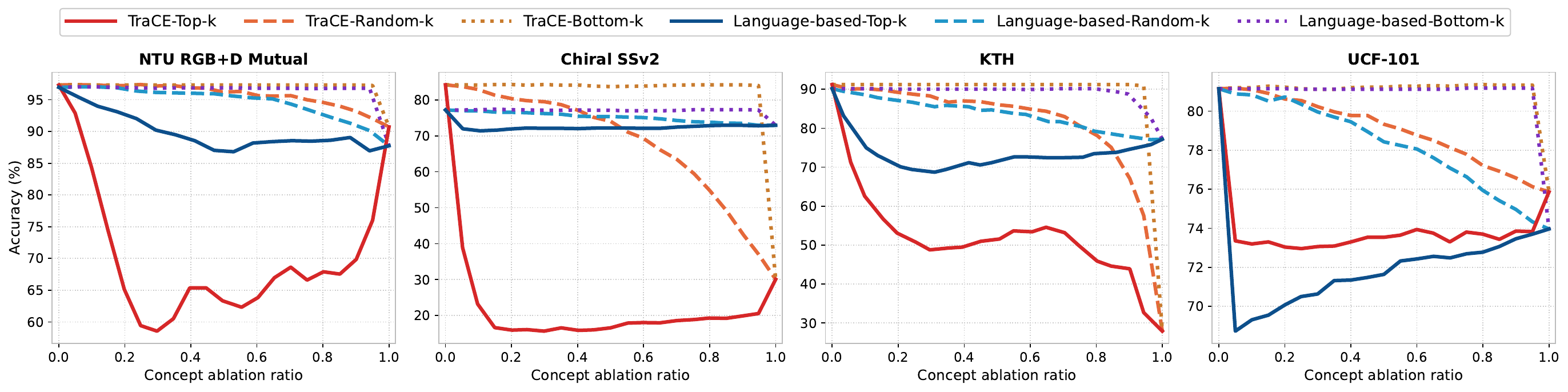}
\vspace{\figcapmargin}
\caption{
\textbf{Concept intervention curves show faithfulness of \ours{}.}
We deactivate concepts in descending (Top-\(k\)), random, and ascending (Bottom-\(k\)) order of contribution and report prediction accuracy as the removal ratio increases.
}
    \label{fig:deletion}
    \vspace{\figmargin}
\end{figure*}

\vspace{\paramargin}

\paragraph{Contribution rankings identify decision-relevant concepts.}
Beyond semantic alignment, we test whether contribution rankings identify concepts used for prediction.
We progressively deactivate trajectory concepts in descending, random, or ascending contribution order.
In \figref{deletion}, top-ranked removal causes sharper accuracy drops than random or bottom-ranked removal on NTU RGB+D Mutual, Chiral SSv2, and KTH; the separation is weaker on UCF-101.
The corresponding native-VQA analysis shows that removing the top 5\% of trajectory concepts reduces accuracy by 11.9 points on TOMATO and 40.5 points on STAR, compared with 4.3 and 1.3 points under random removal (\tabref{native_vqa_deletion}).
These comparisons assess contribution-ranking faithfulness rather than semantic correctness by themselves.

\begin{table*}[t]
\centering
\caption{
\textbf{Count-matched trajectory concept interventions.}
We report accuracy, its change from the unablated model ($\Delta$Acc.), and the correct-to-incorrect flip rate on direction-annotated subsets.
Mean $\pm$ standard deviation is computed over 100 concept-subsampling draws.
See \apref{targeted_interventions} for details.
}
\label{tab:count_matched_interventions}
\small
\setlength{\tabcolsep}{4pt}
\resizebox{0.80\linewidth}{!}{%
\begin{tabular}{lccc@{\hspace{8pt}}ccc}
\toprule
& \multicolumn{3}{c}{Chiral SSv2 (AR)} & \multicolumn{3}{c}{TOMATO (VQA)} \\
\cmidrule(lr){2-4}\cmidrule(lr){5-7}
Removed concepts & Acc. (\%) & $\Delta$Acc. & Flip rate (\%)
                 & Acc. (\%) & $\Delta$Acc. & Flip rate (\%) \\
\midrule
None & 81.1 & -- & -- & 46.4 & -- & -- \\
GT direction & 56.9\textpm0.1 & \textbf{$-$24.2} & \textbf{30.1\textpm0.1}
             & 23.9\textpm0.4 & \textbf{$-$22.5} & \textbf{70.5\textpm0.5} \\
Opposite direction & 82.2\textpm0.1 & +1.1 & 3.6\textpm0.1
                   & 54.0\textpm0.2 & +7.6 & 30.8\textpm0.0 \\
Random non-GT & 82.1\textpm0.2 & +1.0 & 3.7\textpm0.2
              & 45.4\textpm1.3 & $-$1.0 & 21.3\textpm2.0 \\
\bottomrule
\end{tabular}
}
\vspace{\tabmargin}
\end{table*}

\vspace{\paramargin}

\paragraph{Direction-targeted interventions link semantics to predictions.}
To connect directional semantics with decision dependence, 
we use the fixed direction labels above to remove equal numbers of ground-truth (GT)-direction, opposite-direction, or random non-GT concepts on the same directional class subset of Chiral SSv2 used in \tabref{prediction_consistency_avg} and a direction-annotated subset of TOMATO.
As shown in \tabref{count_matched_interventions}, GT-direction removal reduces AR accuracy by 24.2 points and flips 30.1\% of initially correct predictions; neither control reduces accuracy, and both yield flip rates below 4\%.
For VQA, GT-direction removal produces the largest accuracy drop (22.5 points) and flip rate (70.5\%).
The VQA flip rate is descriptive because only 13 of the 28 samples are initially correct. 
These interventions provide model-internal evidence that direction-aligned trajectory concepts influence predictions in both tasks.

\vspace{\paramargin}
\begin{figure}[t]
    \centering
    \includegraphics[width=\linewidth]{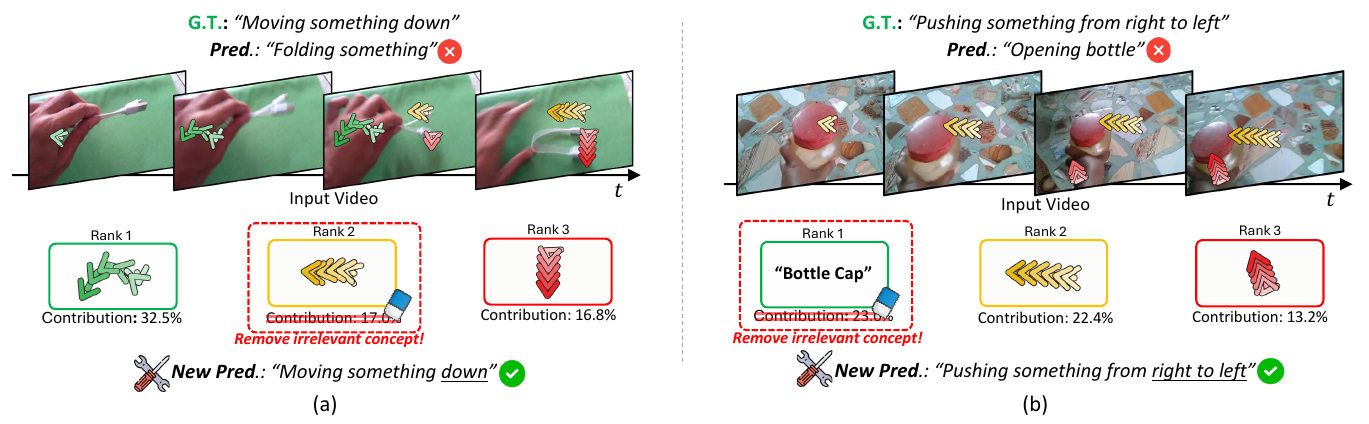}
    \vspace{\figcapmargin}
    \caption{
    \textbf{Concept-level intervention.}
    (a) Deactivating a mismatched trajectory concept fixes a motion error.
    (b) Deactivating the spurious object concept ``bottle cap''
    shifts the model from static context to motion evidence.
    These examples illustrate actionable concept-level debugging with \ours{}.
}
    \label{fig:intervention}
    \vspace{\figmargin}
\end{figure}

\paragraph{Concept intervention supports sample-level debugging.}
These concept-level interventions also enable targeted correction of individual errors.
In \figref{intervention} (a), the model misclassifies \textit{Moving [something] down} as \textit{Folding [something]}, with a mismatched leftward-motion concept contributing to the prediction.
Deactivating that concept corrects the prediction.
In \figref{intervention} (b), the model predicts \textit{Opening bottle} for a \textit{Pushing [something] from right to left} video because the object concept ``bottle cap'' is highly influential.
Removing this spurious object concept shifts the model from static context to motion evidence and corrects the prediction.
\ours{} exposes whether an error stems from misleading motion concepts or static context and enables targeted concept-level correction, illustrating how the concept interface can support model debugging.

\vspace{\secmargin}
\vspace{-0.5em}

\section{Conclusions}

\label{sec:conclusions}

\vspace{\paramargin}
In this work, we address \emph{motion-source under-specification} in explainable video understanding.
We propose \ours{}, a concept-based framework that derives \emph{source-flexible} motion concepts from optical-flow trajectories to capture motion from actors, objects, and interactions while separating dynamic evidence from static object and scene context.
Across action recognition and video question answering, \ours{} maintains competitive task performance while producing explanations that users find interpretable and that intervention analyses show to be faithful to model behavior.
These results suggest that trajectory concepts offer a practical typed interface for explaining motion evidence in video understanding.

\begin{ack}

This work was supported by the National Research Foundation of Korea (NRF) grants funded by the Korea government (MSIT) (RS-2025-02216217, 50\%) and (RS-2026-25616405, 25\%).
Additionally, it was supported in part by the Institute of Information \& Communications Technology Planning \& Evaluation (IITP) grant funded by the Korea Government (MSIT) under grant RS-2024-00353131 (10\%) and the Institute of Information \& Communications Technology Planning \& Evaluation (IITP)-ITRC (Information Technology Research Center) grant funded by the Korea government (MSIT) (IITP-2026-RS-2023-00259004, 10\%).
This research was also supported by the MSIT (Ministry of Science and ICT), Korea, under the National Program for Excellence in SW (2023-0-00042, 5\%) supervised by the IITP (Institute of Information \& Communications Technology Planning \& Evaluation) in 2026.
\end{ack}

{\small
\bibliographystyle{plainnat}
\bibliography{main}
}
\clearpage

\appendix
\section*{Appendix}

In this appendix, we provide additional details on the architecture, the optical flow / trajectory pipeline, datasets, baselines, and the user study, to complement the main paper. We organize the appendix as follows:

\begin{itemize}
    \item Architecture and implementation details (\apref{arch_details})
    \item Dataset details (\apref{dataset})
    \item Baseline details (\apref{baseline})
    \item Setup and protocol of the user study (\apref{userstudy_details})
    \item Additional experimental results and analyses (\apref{add_results})
    \item Limitations (\apref{limitations})
    \item Broader impacts (\apref{broader_impacts})
\end{itemize}

\section{Architecture and Implementation Details}
\label{appen:arch_details}
\label{appen:impl}

In this section, we describe the backbones, the concept layers and task heads, the concept discovery pipeline, and the training details of \ours{}.

\subsection{Action Recognition Backbone}
\label{appen:arch_backbone_ar}
\label{appen:AR_backbone_impl}

For action recognition, we use VideoMAE-B/16~\cite{videomae} as the default video encoder $f_V$.
Given an input video, $f_V$ outputs a video-level feature $\mathbf{x}_i \in \mathbb{R}^{d}$, and the video-side concept layer $g_V$ takes this feature as its input.
VideoMAE-B/16 inherits its layout from ViT-Base ($12$ transformer blocks, hidden dimension $768$, $12$ attention heads, tubelets of size $2 \times 16 \times 16$), and we thus set $d = 768$.

\paragraph{Data preprocessing. }
We sample $T = 16$ frames per video as input to $f_V$.
We use dense sampling for UCF-101~\cite{ucf} and uniform sampling for KTH~\cite{kth}, NTU RGB+D Mutual~\cite{ntu}, and Chiral SSv2~\cite{ssv2,chirality}.
After sampling, we resize each frame so that the shorter side is $224$ pixels and apply random cropping to $224 \times 224$.
We do not apply any other data augmentation; in particular, we disable random erasing, mixup, and CutMix during fine-tuning.

\paragraph{Pretrained weights. }
For Chiral SSv2 and UCF-101, we directly use the publicly released VideoMAE-B/16~\cite{videomae} fine-tuned weights from the official repository, without further fine-tuning.
For KTH and NTU RGB+D Mutual, we initialize $f_V$ from the VideoMAE-B/16 weights pretrained on Something-Something-V2~\cite{ssv2} and fine-tune on each target dataset.
Something-Something-V2 is a large-scale, temporally fine-grained dataset, and the self-supervised VideoMAE weights pretrained on it are particularly effective for downstream datasets where temporal dynamics play a crucial role, such as KTH and NTU RGB+D Mutual.

\paragraph{Training. }
\begin{table}[t]
\centering
\caption{\textbf{Hyperparameters used for fine-tuning the action recognition backbone $f_V$ on KTH and NTU RGB+D Mutual.} For Chiral SSv2 and UCF-101, we directly use the publicly released VideoMAE-B/16 fine-tuned weights from the official repository.}
\label{tab:ar_hparam}
\small
\begin{tabular}{lcc}
\toprule
Config & KTH & NTU RGB+D Mutual \\
\midrule
Model                       & ViT-Base/16          & ViT-Base/16          \\
Pretrained weights          & VideoMAE-B/16 (SSv2) & VideoMAE-B/16 (SSv2) \\
Num.\ frames                & 16                   & 16                   \\
Input size                  & $224 \times 224$     & $224 \times 224$     \\
Short-side size             & 224                  & 224                  \\
Frame sampling              & uniform              & uniform              \\
Optimizer                   & AdamW                & AdamW                \\
Optimizer momentum          & $\beta_1{=}0.9$, $\beta_2{=}0.999$ & $\beta_1{=}0.9$, $\beta_2{=}0.999$ \\
Base learning rate          & $5\text{e}{-4}$      & $1\text{e}{-3}$      \\
Weight decay                & $0.05$               & $0.05$               \\
Learning rate schedule      & cosine decay         & cosine decay         \\
Warmup epochs               & 5                    & 5                    \\
Layer-wise lr decay         & 0.75                 & 0.75                 \\
Training epochs             & 50                   & 50                   \\
GPUs                        & $4 \times$ RTX 3090  & $4 \times$ RTX 3090  \\
Batch size per GPU          & 10                   & 10                   \\
Augmentation                & none                 & none                 \\
Test segments               & 2                    & 2                    \\
Test crops                  & 3                    & 3                    \\
\bottomrule
\end{tabular}
\end{table}

We fine-tune $f_V$ on KTH and NTU RGB+D Mutual for $50$ epochs with $5$ warmup epochs and a cosine learning rate schedule~\cite{cos_decay}.
We use the AdamW optimizer~\cite{adamw} ($\beta_1 = 0.9$, $\beta_2 = 0.999$, weight decay $0.05$), apply layer-wise learning rate decay of $0.75$, and use a base learning rate of $5\text{e}{-4}$ for KTH and $1\text{e}{-3}$ for NTU RGB+D Mutual.
We conduct backbone training on $4$ NVIDIA GeForce RTX 3090 GPUs with a batch size of $10$ per GPU.
We linearly scale the base learning rate with respect to the total batch size using the rule $lr = \text{base learning rate} \times \text{batch size} / 256$, following VideoMAE~\cite{videomae}.
We list the per-dataset hyperparameters in \tabref{ar_hparam}.
After fine-tuning, we freeze $f_V$ and pre-extract its video-level features $\mathbf{x}_i$ for every training and validation sample.
This pre-extraction strategy decouples backbone training from concept-layer training and enables efficient and reproducible optimization of $g_V$ and $\mathbf{W}_{\text{AR}}$.

\subsection{Video Question Answering Backbone}
\label{appen:arch_backbone_qa}
\label{appen:VQA_backbone_impl}

For video question answering, we use the Stage-2 1B variant of InternVideo2~\cite{wang2024internvideo2}---denoted as InternVideo2$_{s2}$-1B---as the video encoder $f_V$, and we use its paired text tower as the text encoder $f_T$.
We adopt InternVideo2$_{s2}$-1B because Stage-2 aligns the video and text encoders into a shared embedding space through video--audio--speech--text contrastive learning.
This alignment matches the video-side concept layer $g_V$ and the text-side concept projector $g_T$ to a shared concept space, rather than to two independently learned spaces, so that the matching operator in \secref{vqa} can compare the two outputs dimension-wise.
The text tower is initialized from the first $19$ layers of BERT-Large~\cite{devlin2019bert}; for each question--answer pair $[Q_i; a_{ij}]$, $f_T$ outputs a text feature $\mathbf{u}_{ij} = f_T([Q_i; a_{ij}]) \in \mathbb{R}^{d_T}$.
We initialize both towers with the publicly released InternVideo2$_{s2}$-1B weights and freeze them throughout VQA training; we only update $g_V$, $g_T$, and the per-group weights $\gamma_g$ used in the matching operator.

\paragraph{Data preprocessing. }
We sample $T = 4$ frames per video using uniform stride sampling: we partition the video into $T$ equal-length segments and select the first frame of each segment.
We resize each frame so that the shorter side is $224$ pixels and apply center cropping to $224 \times 224$.
For text inputs, we tokenize the question--answer pair $[Q_i; a_{ij}]$ with the BERT-Large~\cite{devlin2019bert} tokenizer paired with the InternVideo2 text tower, and we feed the tokenized sequence into the text tower without additional preprocessing.
We do not apply any data augmentation during VQA training.

\paragraph{Non-interpretable VQA baseline. }
The non-interpretable VQA baseline used as the reference point for \ours{} in \tabref{main_results} trains a single linear projection layer on top of the frozen video and text features to adapt them to the multiple-choice format.
Specifically, we apply two separate linear projections, $\phi_V : \mathbb{R}^{d} \rightarrow \mathbb{R}^{d_p}$ and $\phi_T : \mathbb{R}^{d_T} \rightarrow \mathbb{R}^{d_p}$, to the video feature $\mathbf{x}_i = f_V(\mathbf{V}_i)$ and the question--answer feature $u_{ij} = f_T([Q_i; a_{ij}])$, with a shared projection dimension $d_p = 512$.
We $\ell_2$-normalize the two projections and score each candidate answer with their dot product:
\begin{equation}
s_{ij}^{\text{base}} = \exp(\gamma) \cdot \frac{\phi_V(\mathbf{x}_i)}{\|\phi_V(\mathbf{x}_i)\|_2} \cdot \frac{\phi_T(u_{ij})}{\|\phi_T(u_{ij})\|_2},
\end{equation}
where $\gamma$ is a learnable log-temperature that we initialize to $2.3$, so that $\exp(\gamma) \approx 10$ at the start of training.
We optimize the parameters of $\phi_V$, $\phi_T$, and $\gamma$ with the cross-entropy loss over the candidate scores $\{s_{ij}^{\text{base}}\}_{j=1}^{J}$ against the ground-truth answer index $y_i$.
We use the same training configuration across all video question answering datasets (TOMATO~\cite{TOMATO}, STAR~\cite{star}, and the multiple-choice reformulations of NTU RGB+D Mutual~\cite{ntu}, Chiral SSv2~\cite{ssv2,chirality}, and UCF-101~\cite{ucf}); we list the hyperparameters in \tabref{vqa_baseline_hparam}.

\begin{table}[t]
\centering
\caption{\textbf{Hyperparameters for training the baseline projection layer.} The video and text encoders ($f_V$ and $f_T$) are kept frozen; only $\phi_V$, $\phi_T$, and the learnable temperature $\gamma$ are updated. We use the same configuration across all video question answering datasets.}
\label{tab:vqa_baseline_hparam}
\small
\begin{tabular}{lc}
\toprule
Config & Value \\
\midrule
Frozen backbone             & InternVideo2$_{s2}$-1B \\
Num.\ frames                & 4 \\
Input size                  & $224 \times 224$ \\
Frame sampling              & uniform stride \\
Projection dimension $d_p$  & 512 \\
Feature normalization       & $\ell_2$ \\
Temperature init ($\gamma$) & $2.3$ \\
Optimizer                   & Adam~\cite{adam} \\
Base learning rate          & $1\text{e}{-3}$ \\
Weight decay                & $0$ \\
Gradient clipping           & $1.0$ \\
Batch size                  & 256 \\
Training epochs             & 100 \\
Random seed                 & 42 \\
\bottomrule
\end{tabular}
\end{table}

\subsection{Concept Layers and Task Heads}
\label{appenarch_concept_layers}
\label{appenarch_classifier}

\paragraph{Video-side concept layer.}
The video-side concept layer $g_V : \mathbb{R}^{d} \rightarrow \mathbb{R}^{M}$, where $M = M_t + M_o + M_s$, is a group-wise bias-free linear projection with three disjoint weight matrices $\mathbf{W}_V^t \in \mathbb{R}^{M_t \times d}$, $\mathbf{W}_V^o \in \mathbb{R}^{M_o \times d}$, and $\mathbf{W}_V^s \in \mathbb{R}^{M_s \times d}$, giving raw scores $\mathbf{a}_i^g = \mathbf{W}_V^g \mathbf{x}_i$ for $g \in \{t, o, s\}$.
We apply an element-wise sigmoid only to the trajectory branch: $\mathbf{z}_i^t = \sigma(\mathbf{a}_i^t)$, $\mathbf{z}_i^o = \mathbf{a}_i^o$, $\mathbf{z}_i^s = \mathbf{a}_i^s$.
We omit the bias term in all three branches, following prior label-free concept bottleneck models~\cite{lf-cbm}.
We train the trajectory branch with soft-label binary cross-entropy against the soft trajectory presence label $\mathbf{c}_i^t \in [0,1]^{M_t}$ defined in \secref{trajectory_concepts}:
\begin{equation}
\label{eq:loss_t}
L_t = -\frac{1}{M_t} \sum_{k=1}^{M_t} \big[ c_{i,k}^t \log z_{i,k}^t + (1 - c_{i,k}^t) \log(1 - z_{i,k}^t) \big].
\end{equation}
We train the object and scene branches with the cosine-cubed loss~\cite{lf-cbm} against the pseudo-labels $\mathbf{c}_i^g$ defined in \secref{object_scene_concepts}:
\begin{equation}
\label{eq:loss_s}
L_g = -\frac{(\mathbf{z}_i^g)^3 \cdot (\mathbf{c}_i^g)^3}{\|(\mathbf{z}_i^g)^3\|_2 \, \|(\mathbf{c}_i^g)^3\|_2}, \quad g \in \{o, s\}.
\end{equation}
The object and scene branches keep their raw scores because this loss operates on directional alignment rather than per-dimension probabilities.
We instantiate a separate $g_V$ for each task: for action recognition, we freeze $g_V$ after concept bottleneck learning and train only the sparse classifier $\mathbf{W}_{\mathrm{AR}}$; for video question answering, we initialize $g_V$ from the concept bottleneck learning stage and continue updating it under concept supervision.

\paragraph{Text concept layer.}
We use the text-side concept projector $g_T : \mathbb{R}^{d_T} \rightarrow \mathbb{R}^{M}$ only in video question answering.
$g_T$ shares the same group-wise bias-free design as $g_V$, with weight matrices $\mathbf{W}_T^t$, $\mathbf{W}_T^o$, and $\mathbf{W}_T^s$, and produces the question-conditioned concept relevance $\mathbf{r}_{ij} = g_T(\mathbf{u}_{ij}) = [\mathbf{r}_{ij}^t; \mathbf{r}_{ij}^o; \mathbf{r}_{ij}^s]$.
We supervise object and scene relevance by text-concept similarity to natural-language concept names with a candidate-averaged cosine loss:
\begin{equation}
\label{eq:loss_text}
L_{\mathrm{text}}^g = -\frac{1}{J} \sum_{j=1}^{J} \frac{\mathbf{r}_{ij}^g \cdot \mathbf{d}_{ij}^g}{\|\mathbf{r}_{ij}^g\|_2 \, \|\mathbf{d}_{ij}^g\|_2}, \quad g \in \{o, s\},
\end{equation}
where $\mathbf{d}_{ij}^g$ is the text-side pseudo-label obtained by comparing the embedding of $[Q_i; a_{ij}]$ with the concept-name embeddings in $\mathbb{C}^g$, both encoded with the \texttt{all-mpnet-base-v2} model~\cite{wolf2020transformers}.
Trajectory relevance, in contrast, has no language names and cannot be supervised by text-concept similarity.
We therefore align question-conditioned trajectory relevance with video-side trajectory evidence through an InfoNCE loss~\cite{infonce}, where the relevance vector of the correct answer is the positive and other candidates are negatives:
\begin{equation}
\label{eq:loss_align}
L_{\mathrm{align}}^t = -\log \frac{\exp(\mathrm{sim}(\mathbf{z}_i^t, \mathbf{r}_{i y_i}^t) / \tau)}{\sum_{j=1}^{J} \exp(\mathrm{sim}(\mathbf{z}_i^t, \mathbf{r}_{ij}^t) / \tau)},
\end{equation}
where $\mathrm{sim}(\cdot, \cdot)$ is cosine similarity and $\tau$ is a temperature.

\paragraph{Sparse linear classifier for action recognition.}
For action recognition, the sparse linear classifier $\mathbf{W}_{\mathrm{AR}} \in \mathbb{R}^{K \times M}$ predicts $\hat{\mathbf{y}}_i = \mathrm{softmax}(\mathbf{W}_{\mathrm{AR}} \mathbf{z}_i)$, where $K$ is the number of action classes.
We optimize cross-entropy with elastic-net regularization on $\mathbf{W}_{\mathrm{AR}}$~\cite{sparse}:
\begin{equation}
L_{\mathrm{AR}} = -\log \hat{\mathbf{y}}_i[y_i] \;+\; \lambda \big[ (1-\alpha)\tfrac{1}{2} \|\mathbf{W}_{\mathrm{AR}}\|_F^2 + \alpha \|\mathbf{W}_{\mathrm{AR}}\|_{1,1} \big],
\end{equation}
where $\|\cdot\|_F$ is the Frobenius norm and $\|\cdot\|_{1,1}$ is the element-wise $\ell_1$ norm.
We optimize $\mathbf{W}_{\mathrm{AR}}$ with the GLM-SAGA solver~\cite{sparse}, following prior label-free concept bottleneck models~\cite{lf-cbm,dance}.

\paragraph{Sparse matching operator for video question answering.}
For video question answering, we score each candidate answer with the sparse matching operator in \secref{vqa}.
We use $K_t = K_o = K_s = 5$ across all video question answering experiments and set $\epsilon = 10^{-8}$ in the ReLU $\ell_2$-normalization $\nu(\cdot)$.
We initialize $\gamma_g = 0$ for all groups, which corresponds to $\exp(\gamma_g) = 1$ at the start of training, so that all three concept groups contribute equally, and we update $\gamma_g$ jointly with $g_V$ and $g_T$.

\subsection{Concept Discovery}
\label{appen:concept_discovery}

\subsubsection{Trajectory Concept}
\label{appen:trajectory_concept}

\paragraph{Optical flow estimation. }
We use WAFT~\cite{waft} as the optical flow estimator across all experiments.
WAFT is a pretrained dense flow estimator that produces a per-pixel displacement field between two consecutive frames.
We use the official released WAFT weights and do not fine-tune the estimator on any target dataset.
Keeping the flow estimator frozen makes the trajectory pipeline backbone-agnostic and reproducible across datasets.
For each video, we sample frames at the original frame rate and compute forward flow between every pair of adjacent frames.
We perform optical flow estimation once per dataset and cache the resulting flow fields and trajectories on disk, so that trajectory clustering and concept-layer training can both reuse the cached results without recomputation.

\paragraph{Trajectory generation. }
Given a video $\mathbf{V}_i$, we partition it into $N_s$ temporal segments of equal duration and extract an $L$-frame clip centered at each segment center.
Within a segment, let $\mathbf{F}_r : \Omega \rightarrow \mathbb{R}^{2}$ denote the forward flow field from frame $r$ to frame $r+1$, where $\Omega = \{0, \ldots, W-1\} \times \{0, \ldots, H-1\}$ is the discrete pixel grid.
For each seed pixel $\mathbf{p}_0 \in \Omega$, we trace a visual trajectory by recursively accumulating optical flow:
\begin{equation}
\mathbf{p}_{r+1} = \mathbf{p}_r + \mathbf{F}_r(\mathbf{p}_r), \quad r = 0, \ldots, L-2.
\end{equation}
Although the seed $\mathbf{p}_0$ lies on the discrete grid, subsequent positions $\mathbf{p}_r$ may take sub-pixel coordinates.
We therefore evaluate $\mathbf{F}_r(\mathbf{p}_r)$ via bilinear interpolation.
We represent each trajectory as the concatenated sequence of displacement vectors,
\begin{equation}
\boldsymbol{\tau}(\mathbf{p}_0) = [\Delta \mathbf{p}_0, \ldots, \Delta \mathbf{p}_{L-2}] \in \mathbb{R}^{2(L-1)},
\end{equation}
where $\Delta \mathbf{p}_r = \mathbf{p}_{r+1} - \mathbf{p}_r$.
Applying this procedure to every segment yields the raw trajectory set $\widetilde{\mathbb{T}}_i$ for $\mathbf{V}_i$.
The dataset-specific values of $L$ and $N_s$ are listed in \apref{trace}.

\paragraph{Trajectory filtering. }
Raw dense trajectories contain three failure modes: static background, weak or near-zero motion, and noisy flow predictions in low-texture regions.
We address all three with a two-stage filter applied per segment.
For notational simplicity, we describe the procedure for one segment and omit the segment index.
Let $\widetilde{\mathcal{P}}$ denote the set of seed pixels in the segment, and let $M : \Omega \rightarrow \{0, 1\}$ denote the foreground saliency mask obtained by an off-the-shelf foreground estimator~\cite{u2net} on the segment's center frame.
\textbf{Stage 1 (foreground gating).} For each seed pixel $\mathbf{p}_0 \in \widetilde{\mathcal{P}}$, we retain its trajectory only if $\mathbf{M}(\mathbf{p}_0) = 1$.
This focuses the trajectory pool on visible entity-level motion such as people, manipulated objects, and interacting agents.
\textbf{Stage 2 (magnitude ranking).} Among the remaining foreground trajectories, we compute the motion magnitude
\begin{equation}
m(\boldsymbol{\tau}) = \sum_{r=0}^{L-2} \|\Delta \mathbf{p}_r\|_2,
\end{equation}
and retain the top-$\rho\%$ trajectories ranked by $m(\boldsymbol{\tau})$.
We use $\rho = 5$ in all experiments.
This stage suppresses low-amplitude flow noise and concentrates the trajectory vocabulary on salient entity-level motion.
Applying both stages to every segment yields the filtered trajectory set $\mathbb{T}_i$ for $\mathbf{V}_i$.

\paragraph{Trajectory clustering and labeling. }
We aggregate the filtered trajectories over the training set, $\mathbb{T} = \bigcup_{i=1}^{N} \mathbb{T}_i$, and cluster them with FINCH~\cite{finch}.
FINCH produces a hierarchy of partitions at different levels of granularity without requiring a predefined number of clusters.
We select a single partition level for each downstream dataset, and the resulting number of clusters $M_t$ varies across datasets accordingly.
We list the partition level and the resulting $M_t$ per dataset in \apref{trace}.
Let $\pi : \mathbb{T} \rightarrow \{1, \ldots, M_t\}$ denote the cluster assignment of the selected partition.
The $k$-th trajectory concept is the set of trajectories assigned to cluster $k$, $C_k^{t} = \{\boldsymbol{\tau} \in \mathbb{T} \mid \pi(\boldsymbol{\tau}) = k\}$.
Each concept captures a recurring trajectory-level motion pattern shared across videos.
For each video $\mathbf{V}_i$, we count how often its filtered trajectories are assigned to each concept,
\begin{equation}
h_{i,k} = \sum_{\boldsymbol{\tau} \in \mathbb{T}_i} \mathbbm{1}[\pi(\boldsymbol{\tau})=k],
\end{equation}
and form a soft trajectory presence label
\begin{equation}
\mathbf{c}_i^t = \frac{\mathbf{h}_i}{\|\mathbf{h}_i\|_1} \in [0, 1]^{M_t}.
\end{equation}
The entry $c_{i,k}^t$ is the relative frequency of trajectory concept $k$ in $\mathbf{V}_i$, and serves as the soft target for the trajectory branch in Equation~\eqref{eq:loss_t}.

\paragraph{Concept visualization. }
To visualize a trajectory concept, we select the medoid trajectory of its cluster as the canonical example.
The medoid is the trajectory with the smallest average distance to all other members of the cluster, measured in the same flattened displacement space used for clustering.
For each visualized concept, we identify the most strongly activating training video and locate the region in that video where the concept activation comes from using Grad-CAM~\cite{gradcam}.
We then overlay the medoid trajectory on the Grad-CAM-localized region, so that the visualization shows both the canonical motion pattern and the spatial region in the video that triggers the concept.
We use Grad-CAM only to localize concept responses for visualization purposes, not to compute the explanations themselves.
We follow this same protocol for all qualitative figures involving trajectory concepts.

\begin{figure*}[t]
\includegraphics[width=\textwidth]{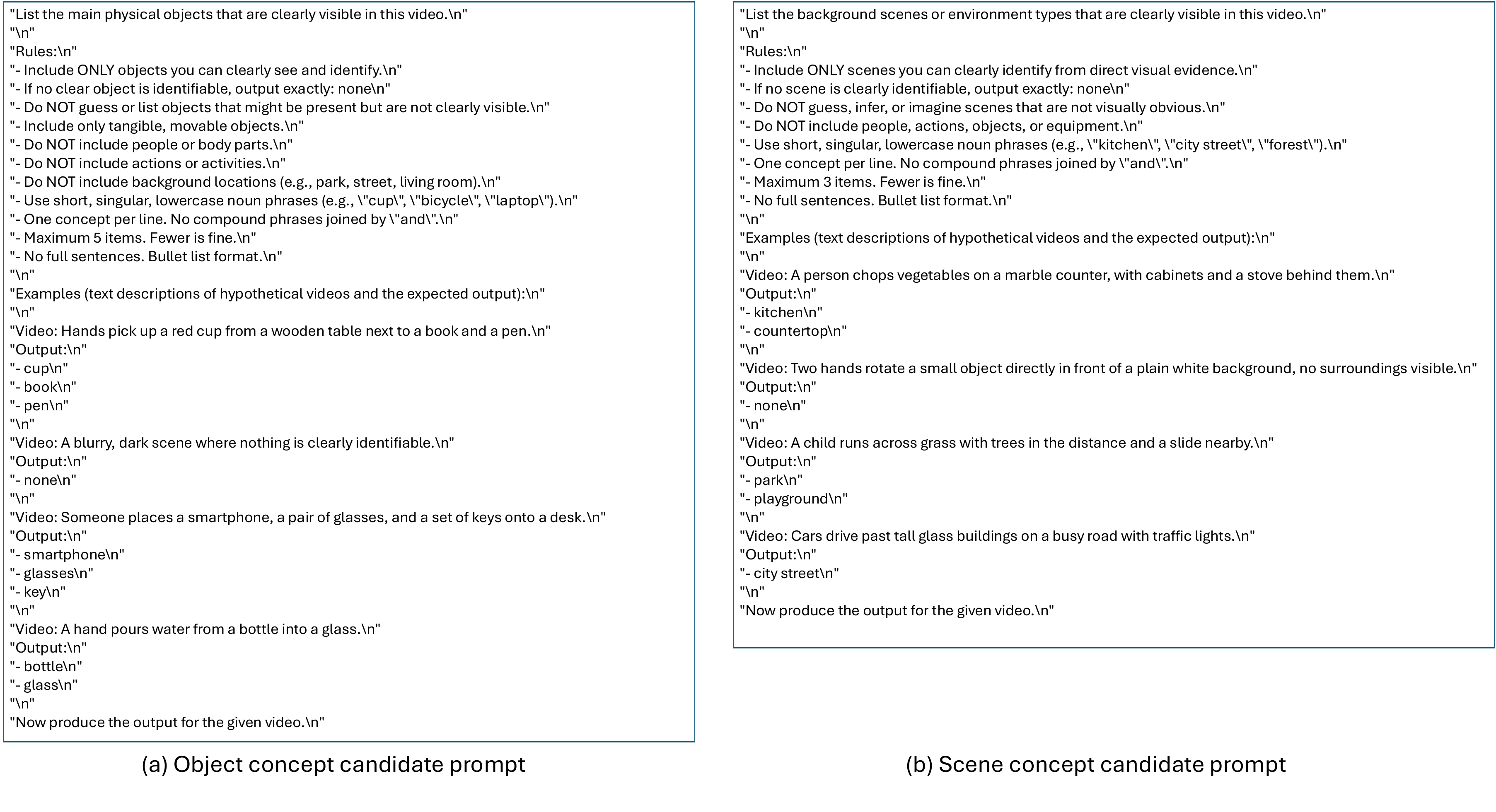}
\caption{\textbf{Concept candidate generation prompts.} (a) Object and (b) scene prompts used to query Qwen3-VL~\cite{qwen3vl}.}
\label{fig:prompt}
\end{figure*}

\subsubsection{Object and Scene Concept}
\label{appen:object_and_scene_concept}

\paragraph{Object and scene concept candidate generation.}
We construct object and scene concept vocabularies by querying the training videos with the Video-LLM Qwen3-VL~\cite{qwen3vl}.
Unlike prior class-name prompting~\cite{lf-cbm,dance}, this video-grounded querying makes the candidate vocabulary reflect visible evidence rather than action-category priors.
For each training video, we run two independent Video-LLM queries, one for object candidates and one for scene candidates, and we union the per-video outputs over the training set.
The object prompt asks Qwen3-VL to list the main physical objects clearly visible in the video, with explicit instructions to exclude people, body parts, actions, and background locations, and to output a short bullet list of singular lowercase noun phrases.
The scene prompt asks Qwen3-VL to list the background scenes or environment types clearly visible in the video, with explicit instructions to exclude people, actions, objects, and equipment.
Both prompts cap the output length and provide in-context examples to enforce the bullet-list format.
We provide the full prompt templates in \figref{prompt}.
Aggregating the per-video outputs yields the candidate sets $\widetilde{\mathbb{C}}^{o}$ and $\widetilde{\mathbb{C}}^{s}$.

\paragraph{Object and scene concept filtering. }
We refine $\widetilde{\mathbb{C}}^{o}$ and $\widetilde{\mathbb{C}}^{s}$ with four filters from LF-CBM~\cite{lf-cbm}.
\textbf{(i) Length filter.} We discard any candidate concept longer than 30 characters to keep concepts simple and to avoid unnecessarily long phrases.
\textbf{(ii) Redundancy filter.} We remove duplicate and synonymous concepts using a weighted text-embedding similarity score that combines two encoders: MPNet~\cite{wolf2020transformers} and CLIP ViT-B/16~\cite{clip}.
\textbf{(iii) Activation filter.} For each remaining candidate, we compute its average video--text similarity score under the pretrained vision--language encoder~\cite{wang2024internvideo2} across the training set, and we discard concepts whose average score falls below an activation cutoff.
This filter removes concepts that are not actually represented in the training data.
\textbf{(iv) Projection filter.} After training the concept layer with the surviving candidates, we further remove concepts whose projected activations are consistently low across the training set, using a projection cutoff.
This filter removes concepts that the model does not rely on after optimization.
We do not apply the LF-CBM filter that removes concepts similar to target class names, because our object and scene candidates are generated from video evidence rather than action class names and rarely overlap with class labels.
We tune the activation and projection cutoffs per dataset; the exact values and the resulting vocabulary sizes $M_o$ and $M_s$ are reported in \apref{trace}.

\paragraph{Concept labeling.}
For each training video $\mathbf{V}_i$ and each concept group $g \in \{o, s\}$, we generate a soft pseudo-label using a pretrained vision-language encoder $E_T(\cdot), E_V(\cdot)$.
We use InternVideo2~\cite{wang2024internvideo2} as the encoder to align the pseudo-labels with the VQA backbone described in \apref{VQA_backbone_impl}.
Specifically, we encode every concept name in $\mathbb{C}^{g}$ with the text encoder, yielding the $\ell_2$-normalized concept embedding matrix $E_T(\mathbb{C}^{g}) \in \mathbb{R}^{M_g \times d_E}$.
We encode the video $\mathbf{V}_i$ with the video encoder, yielding $E_V(\mathbf{V}_i) \in \mathbb{R}^{d_E}$.
The soft pseudo-label is
\begin{equation}
\mathbf{c}_i^{g} = E_T(\mathbb{C}^{g}) \, E_V(\mathbf{V}_i) \in \mathbb{R}^{M_g}, \quad g \in \{o, s\}.
\end{equation}
We use the raw cosine similarity scores as soft pseudo-labels without further normalization.
We use $\mathbf{c}_i^{o}$ and $\mathbf{c}_i^{s}$ as the soft targets for the object and scene branches in Equation~\eqref{eq:loss_s}.

\subsection{\ours{} Training Details}
\label{appen:trace}

After training (or selecting) the backbone encoder, we freeze its parameters and pre-extract feature vectors for all training and validation samples.
We conduct all concept-layer and classifier training on a single NVIDIA GeForce RTX 3090 GPU.

\paragraph{Video-side concept layer training. }
We train the video-side trajectory branch $\mathbf{W}_V^t$ separately from the object and scene branches $\mathbf{W}_V^o$ and $\mathbf{W}_V^s$, following DANCE~\cite{dance}.
The trajectory branch optimizes $L_t$ in \eqnref{loss_t} against the soft trajectory presence label $\mathbf{c}_i^t$ obtained from FINCH cluster assignments (\apref{trajectory_concept}), and the object and scene branches optimize $L_g$ in \eqnref{loss_s} against the InternVideo2 pseudo-labels $\mathbf{c}_i^g$ (\apref{object_and_scene_concept}).
For all branches, we use the Adam optimizer~\cite{adam} with a learning rate of $1\text{e}{-3}$ and a batch size of $256$, and we train for $3{,}000$ steps with early stopping based on validation loss.
We use the same hyperparameters across all action recognition and video question answering datasets.

\paragraph{Text-side trajectory concept layer training. }
We train the text-side trajectory branch $\mathbf{W}_T^t$ jointly with the video-side trajectory branch $\mathbf{W}_V^t$ and the trajectory matching weight $\gamma_t$, since trajectory concepts have no natural-language names and the text branch must instead be supervised by the video-side trajectory evidence (\secref{vqa}).
We retain the video-side trajectory concept loss $L_t$ during VQA training and optimize it together with the trajectory alignment loss $L_{\text{align}}^t$ in \eqnref{loss_align} and the answer cross-entropy $L_{\text{QA}}$, following \eqnref{loss_vqa}.
We use the Adam optimizer with a learning rate of $1\text{e}{-3}$ and a batch size of $256$, and we train for $100$ epochs.
We initialize $\mathbf{W}_V^t$ from concept bottleneck learning, initialize $\mathbf{W}_T^t$ from scratch, and set $\gamma_t = 0$.

\paragraph{Text-side object and scene concept layer training. }
During VQA training, we jointly update the text-side object and scene branches $\mathbf{W}_T^o, \mathbf{W}_T^s$, the corresponding video-side branches, and their matching weights $\gamma_o, \gamma_s$.
The objective combines the cosine-cubed loss on the video side ($L_o + L_s$ in \eqnref{loss_s}), the candidate-averaged cosine loss on the text side ($L_{\text{text}}^o + L_{\text{text}}^s$ in \eqnref{loss_text}), and the video question answering cross-entropy $L_{\text{QA}}$, with weights of $0.8$ on the concept losses and $0.2$ on $L_{\text{QA}}$.
We use the Adam optimizer with a learning rate of $1\text{e}{-3}$ and a batch size of $256$, and we train for $5{,}000$ steps.
We initialize $\mathbf{W}_V^o$ and $\mathbf{W}_V^s$ from concept bottleneck learning, use Xavier uniform initialization for $\mathbf{W}_T^o$ and $\mathbf{W}_T^s$, and set $\gamma_o = \gamma_s = 0$.

\paragraph{Overall VQA objective. }
The overall VQA objective follows \eqnref{loss_vqa}:
\begin{equation}
L_{\text{VQA}} = \lambda_{\text{QA}} L_{\text{QA}} + \lambda_{\text{concept}} L_{\text{concept}} + \lambda_{\text{text}} (L_{\text{text}}^{o} + L_{\text{text}}^{s}) + \lambda_{\text{align}} L_{\text{align}}^t,
\end{equation}
with $\lambda_{\text{QA}} = 0.2$, $\lambda_{\text{concept}} = 0.8$, $\lambda_{\text{text}} = 0.8$, and $\lambda_{\text{align}} = 1.0$, shared across all video question answering datasets.

\paragraph{Interpretable classifier for action recognition training. }
We employ the GLM-SAGA solver~\cite{sparse} with $\alpha = 0.99$, $\lambda = 0.05$, a step size of $5{,}000$, and a batch size of $512$, following DANCE~\cite{dance}.
Early stopping is applied based on validation loss to avoid overfitting.
During this stage, the video encoder $f_V$ and the video-side concept layer $g_V$ are both frozen; only $\mathbf{W}_{\text{AR}}$ is updated.

\paragraph{Concept configuration. }
\begin{table}[t]
\centering
\caption{\textbf{Dataset-specific concept configurations used by \ours{}.} We report the trajectory length $L$, the number of segments per video $N_s$, the number of trajectory concepts $M_t$ after FINCH clustering, the activation and projection cutoffs used to filter object and scene candidates, and the number of object and scene concepts after / before filtering ($M_o$ and $M_s$). Cutoffs marked ``--'' are not applied: TOMATO and STAR are video question answering datasets evaluated with the unfiltered object and scene vocabularies. Trajectory concepts have no candidate filtering stage.}
\label{tab:concept_config}
\small
\setlength{\tabcolsep}{4pt}
\resizebox{\linewidth}{!}{
\begin{tabular}{lccccccc}
\toprule
Dataset & $L$ & $N_s$ & $M_t$ & Activation cutoff & Projection cutoff & $M_o$ (after\,/\,before) & $M_s$ (after\,/\,before) \\
\midrule
NTU RGB+D Mutual & 16 & 5  & 121 & 0.3 & 0.2 & 55\,/\,80     & 28\,/\,28 \\
Chiral SSv2      & 16 & 3  & 154 & 0.3 & 0.2 & 404\,/\,1093  & 286\,/\,519 \\
UCF-101          & 16 & 3  & 134 & 0.3 & 0.3 & 1103\,/\,1181 & 578\,/\,615 \\
KTH              & 16 & 6  & 111 & 0.1 & 0.1 & 26\,/\,27     & 33\,/\,33 \\
TOMATO & 16 & 23 & 70  & -- & -- & 182\,/\,182 & 94\,/\,94   \\
STAR   & 50 & 16 & 274 & -- & -- & 513\,/\,513 & 142\,/\,142 \\
\bottomrule
\end{tabular}
}
\end{table}

We describe the dataset-specific concept configurations used across all experiments in \tabref{concept_config}.
For each dataset, we list the trajectory length $L$, the number of segments per video $N_s$, the resulting number of trajectory concepts $M_t$ produced by FINCH~\cite{finch}, the activation and projection cutoffs used to filter object and scene candidates, and the number of object and scene concepts before and after filtering, $M_o$ and $M_s$.
The trajectory length $L$ is set to $16$ for all datasets except STAR~\cite{star}, where we increase it to $50$ to capture longer-horizon motion required for situated reasoning.
We do not apply the activation or projection cutoffs to TOMATO and STAR, since these video question answering datasets are evaluated with a fixed object and scene vocabulary that we keep unfiltered to preserve coverage over diverse question contents.
Trajectory concepts have no candidate filtering stage; their count $M_t$ is determined entirely by FINCH's automatic partition selection on the filtered trajectory set $\mathbb{T}$ (\apref{trajectory_concept}).

\paragraph{Inference. }
During inference, we resize each frame so that the shorter side is $224$ pixels and apply a center crop of $224 \times 224$.
We do not apply any test-time augmentation and report all results using a single-view setting ($1$ clip $\times$ $1$ crop).
For action recognition, the predicted class is $\hat{c}_i = \arg\max_c [\mathbf{W}_{\text{AR}} \mathbf{z}_i]_c$.
For video question answering, the predicted answer is $\hat{a}_i = a_{ij_i^{\text{pred}}}$, where $j_i^{\text{pred}} = \arg\max_j s_{ij}$.

\section{Dataset Details}
\label{appen:dataset}

We evaluate \ours{} on a suite of action recognition and video question answering datasets.
The action recognition datasets test whether trajectory concepts capture dynamic evidence across diverse motion sources, while the video question answering datasets test whether the same typed concept interface supports language-grounded reasoning.
 
\subsection{Action Recognition Datasets}
\label{appen:ar_datasets}
 
\paragraph{NTU RGB+D. }
NTU RGB+D~\cite{ntu} is a large-scale indoor action recognition dataset captured with multiple camera views and Kinect depth sensors.
We use two subsets: NTU RGB+D Mutual and NTU RGB+D Single.
NTU RGB+D Mutual consists of $5$ mutual-interaction classes where two people interact with each other: \emph{pushing other person}, \emph{hugging other person}, \emph{giving something to other person}, \emph{walking towards each other}, and \emph{walking apart from each other}.
NTU RGB+D Single consists of $12$ single-person classes organized into $6$ opposing pairs (e.g., \emph{drop} vs.\ \emph{pickup}, \emph{sitting down} vs.\ \emph{standing up}, \emph{wear jacket} vs.\ \emph{take off jacket}).
We use the standard cross-subject split for both subsets.
We include NTU RGB+D Mutual as the primary benchmark in \tabref{main_results} of the main paper because multi-person interactions directly test whether trajectory concepts can explain motion from multiple interacting agents beyond single-person pose. We use NTU RGB+D Single only in the concept-quality user study (Task~2 in \apref{userstudy_details}). 
 
\paragraph{Chiral SSv2. }
Chiral SSv2~\cite{ssv2,chirality} is a subset of Something-Something V2~\cite{ssv2}, consisting of $32$ classes where paired classes differ by signed motion direction (e.g., \emph{pulling [something] left to right} vs.\ \emph{right to left}).
This design makes the dataset a strong test of whether trajectory concepts track motion direction rather than static appearance.
We use the official Chiral SSv2 split.
 
\paragraph{UCF-101. }
UCF-101~\cite{ucf} consists of $101$ human action classes collected from YouTube.
Many classes can be recognized from static scene or object cues, making it a useful test of whether trajectory concepts provide complementary dynamic evidence alongside object and scene concepts.
We use the official split 1 for all experiments.
 
\paragraph{KTH. }
The KTH Action dataset~\cite{kth} consists of $6$ action classes performed by $25$ subjects in $4$ different scenarios.
All actions involve single-person body motion (walking, jogging, running, boxing, hand waving, and hand clapping), making KTH a controlled test of basic trajectory concept quality on well-separated temporal patterns.
We use the standard split.
 
\subsection{Video Question Answering Datasets}
\label{appen:vqa_datasets}
 
\paragraph{TOMATO. }
TOMATO~\cite{TOMATO} is a temporal reasoning benchmark for multimodal foundation models that requires understanding of temporal order, duration, and frequency.
The dataset provides human-centric, real-world, and simulated video scenarios.
Since no official training split is provided, we use the human-centric and simulated scenarios as the training set and the real-world scenarios as the test set.
Each question is a multiple-choice item with four answer candidates.
 
\paragraph{STAR. }
STAR~\cite{star} is a situated reasoning benchmark built upon real-world videos of human actions and interactions.
The dataset includes four question types: interaction, sequence, prediction, and feasibility, each requiring structured situation comprehension and logical reasoning.
We use the official train/val/test split.
Each question is a multiple-choice item with four answer candidates.
 
\paragraph{Action Recognition-to-QA Reformulation. }
To test whether the same typed concept interface supports both class prediction and language-grounded reasoning, we reformulate three action recognition datasets (NTU RGB+D Mutual, Chiral SSv2, and UCF-101) into a multiple-choice video question answering format.
Given a video, we pose the question ``What is the action shown in this video?'' and use all action class names as answer candidates.
The ground-truth answer is the original action label.
This reformulation lets us evaluate \ours{}'s VQA instantiation on the same videos used for action recognition, enabling a direct comparison of the two task heads under the same typed concept interface.
 
\subsection{Generalization Settings}
\label{appen:generalization}
 
\paragraph{Object-disjoint Split of Chiral SSv2. }
To evaluate whether trajectory concepts capture motion patterns rather than object appearance shortcuts, we construct an object-disjoint split of Chiral SSv2 using the Something-Else object annotations~\cite{somethingelse}.
We partition the training and test sets so that the object categories in the two sets do not overlap.
This forces the model to rely on motion evidence from trajectory concepts rather than recognizing familiar objects.
We report results in \tabref{generalizability} of the main paper.
 
\paragraph{Cross-domain. }
We evaluate cross-domain transfer on $65$ TOMATO~\cite{TOMATO} videos labeled with leftward, rightward, upward, or downward motion.
We directly apply the trajectory-only \ours{} pipeline trained on Chiral SSv2~\cite{ssv2,chirality}, keeping the video encoder, concept vocabulary, concept layer, and classification head fixed.
For each video, we select the top-1 prediction from the original $32$ Chiral SSv2 classes.
A prediction is correct if it matches a source action class corresponding to the video's direction.
For leftward motion, we accept \textit{Pushing [something] from right to left} or \textit{Pulling [something] from right to left}.
For rightward motion, we accept \textit{Pushing [something] from left to right} or \textit{Pulling [something] from left to right}.
For upward motion, we accept \textit{Moving [something] up}.
For downward motion, we accept \textit{Moving [something] down}.
All other predictions count as incorrect.
We use the same evaluation criterion for the non-interpretable baseline.

\section{Baseline Details}
\label{appen:baseline}

We provide implementation details for every baseline compared against \ours{} in the main paper.
For fair comparison, all baselines share the same backbone encoder as \ours{} unless explicitly noted.

\subsection{Saliency-based Explanations}
\label{appen:baseline_saliency}

\paragraph{GradCAM.}
Grad-CAM~\cite{gradcam} generates class-discriminative saliency maps by computing the gradient of the predicted class score with respect to intermediate layer activations.
We use Grad-CAM solely for visualization purposes: to localize which spatial region in a video activates a given trajectory concept (\apref{trajectory_concept}).
For the VideoMAE-B/16 backbone~\cite{videomae}, we extract Grad-CAM maps from \texttt{blocks[9].norm}; for the InternVideo2 video tower~\cite{wang2024internvideo2}, we use \texttt{blocks[38].norm}.
Grad-CAM is not used as a baseline explanation method in any quantitative comparison.

\paragraph{SALF-CBM. }
SALF-CBM~\cite{salf-cbm} is a spatially-aware concept bottleneck model originally designed for image classification, generating visual explanations via spatially localized concept activations.
We extend it to video by adding a temporal axis to the spatial concept maps while keeping the rest of the architecture unchanged.
We use the same frozen VideoMAE-B/16 backbone and follow the official implementation for all other components.
We compare against SALF-CBM in the user study (\figref{us1} of the main paper) and in \tabref{main_results} for action recognition.

\subsection{Concept-based Explanations}
\label{appen:baseline_concept}

\paragraph{LF-CBM. }
Label-Free CBM~\cite{lf-cbm} discovers concept vocabularies from language descriptions and trains a concept bottleneck model without manual concept annotation.
We use the official LF-CBM codebase without modification, replacing only the backbone with the same frozen VideoMAE-B/16 features used by \ours{}.
LF-CBM uses language-derived concepts for all concept types, including motion, which limits its ability to represent fine-grained motion patterns that are difficult to express in words.
We compare against LF-CBM in \tabref{main_results} for action recognition.

\paragraph{VLG-CBM. }
VLG-CBM~\cite{vlgcbm} trains concept bottleneck models with vision-language guidance, using CLIP-based concept scoring to improve concept quality.
We do not compare against VLG-CBM as a baseline in \tabref{main_results}.
Instead, we adopt the Number of Effective Concepts (NEC) metric proposed by VLG-CBM to evaluate concept quality and report results in the additional experimental results (\apref{app_analysis}).

\paragraph{DANCE. }
DANCE~\cite{dance} is a concept-based video explanation method that represents motion using human pose sequences extracted by a pretrained pose estimator.
We use the same frozen VideoMAE-B/16 backbone and follow the official implementation.
DANCE defines motion concepts over single-person pose sequences, which makes its motion vocabulary source-incomplete: it cannot directly represent object trajectories and multi-agent interactions.
We compare against DANCE in \tabref{main_results} for action recognition and in the user study (\figref{us1} of the main paper).

\subsection{Language-based Explanations}
\label{appen:baseline_language}
 
\paragraph{GPT-5.4-derived concepts. }
We construct a language-based motion concept baseline by prompting GPT-5.4~\cite{gpt5} to generate motion-describing concepts for each action class.
We then use these language-derived motion concepts in the same concept bottleneck structure as \ours{}, replacing trajectory concepts with the generated language motion concepts while keeping the same object and scene concepts, concept layer architecture, and sparse classifier.
This baseline tests whether language descriptions can substitute for visual trajectory concepts in capturing fine-grained motion evidence.
We compare against this baseline in the user study (\figref{us1} of the main paper) and in the concept intervention curves (\figref{deletion} of the main paper).
GPT-5.4 is used only offline to generate the concept vocabulary, not as a video classifier.
The unablated accuracies of this baseline at the zero-removal point of the curves in \figref{deletion} are $97.0\%$, $77.2\%$, $88.2\%$, and $90.2\%$ on NTU RGB+D Mutual, Chiral SSv2, UCF-101, and KTH, respectively, compared with $97.5\%$, $84.2\%$, $88.6\%$, and $92.0\%$ for \ours{}; its stronger deletion effect on UCF-101 therefore does not result from a higher initial accuracy.

\section{User Study Setup and Protocol}
\label{appen:userstudy_details}

To assess the interpretability of \ours{}, we conducted a user study to evaluate the quality of trajectory concepts and the effectiveness of trajectory-based explanations.
The results of this study are summarized in \figref{us1} of the main paper, and we provide further details in this section.
Specifically, our study aimed to answer two key questions:
 
\begin{itemize}
\item Does \ours{}'s explanation better help users understand the model's prediction compared to other explanation methods?
\item Do trajectory concepts effectively convey motion patterns in the video?
\end{itemize}
 
\subsection{Recruitment}
\label{appen:recruitment}
 
We collected responses from $50$ adult participants recruited through the Prolific platform from English-speaking regions, regardless of gender.
Each participant answered $57$ questions, which included $2$ sanity check questions to ensure engagement and attention.
Participants were compensated \pounds$9$ for the entire study.
On average, participants completed the study in about $27$ minutes.
This study was reviewed and approved as exempt by the Institutional Review Board (IRB) at our institution.
We did not collect any personal information from the participants, nor did we disclose any details about the model or research objectives during the study.
 
\subsection{Study Design}
\label{appen:study_design}
 
\subsubsection{Task 1: Quality of \ours{}'s Explanation}
\label{appen:task1}
 
Task 1 aims to address the first key question: \emph{Does \ours{}'s explanation better help users understand the model's prediction compared to other explanation methods?}
To evaluate the interpretability of \ours{}, we compare its explanations with those of three baseline methods: DANCE~\cite{dance}, which represents motion using human pose sequences, a language-based CBM using GPT-5.4-derived motion concepts~\cite{gpt5}, and SALF-CBM~\cite{salf-cbm}, a saliency-based concept bottleneck model.
 
\paragraph{Experimental setting. }
We conduct pairwise comparisons on motion-focused subsets of NTU RGB+D Mutual~\cite{ntu} and Chiral SSv2~\cite{ssv2,chirality}.
For each comparison, participants are shown a video and two explanations (Explanation A and Explanation B), where one is from \ours{} and the other is from a baseline.
The assignment of \ours{} to A or B is randomized.
Each explanation shows the top-3 contributing concepts overlaid on the input video.
 
\paragraph{Evaluation protocol. }
Participants are asked: ``Which explanation better helps you understand why the model predicted the action?''
Responses are collected on a five-point scale: \emph{A is much better}, \emph{A is slightly better}, \emph{About the same}, \emph{B is slightly better}, and \emph{B is much better}.
Each participant answers $18$ pairwise comparison questions across the three baselines.
We report the distribution of responses in \figref{us1} (a) of the main paper.
\subsubsection{Task 2: Quality of \ours{}'s Trajectory Concept}
\label{appen:task2}
 
Task 2 aims to address the second key question: \emph{Do trajectory concepts effectively convey motion patterns in the video?}
This task evaluates the interpretability of individual concepts by asking whether the concept matches the motion observed across its most strongly activating video clips.
 
\paragraph{Experimental setting. }
We compare three types of concepts: (i) \emph{Trajectory concepts} from \ours{}, (ii) \emph{Pose sequence concepts} from DANCE~\cite{dance}, and (iii) \emph{Language-derived motion concepts} generated by GPT-5.4~\cite{gpt5}.
For each concept type, we randomly select one concept and visualize it with its top-4 most strongly activating video clips.
This task is conducted on motion-focused subsets of NTU RGB+D Mutual, NTU RGB+D Single, and Chiral SSv2.
 
\paragraph{Evaluation protocol. }
Participants are asked: ``How well does the given concept match the actions or motions shown in the video above?''
Responses are collected on a five-point Likert scale (1: Strongly Disagree to 5: Strongly Agree).
We independently evaluate each concept type without directly comparing them to each other.
Each participant answers $37$ concept evaluation questions across the three concept types.
We report the average Likert scores in \figref{us1} (b) of the main paper.
 
\subsection{Interface and Instructions}
\label{appen:interface}
\begin{figure}[h]
\centering
\includegraphics[width=1.0\textwidth]{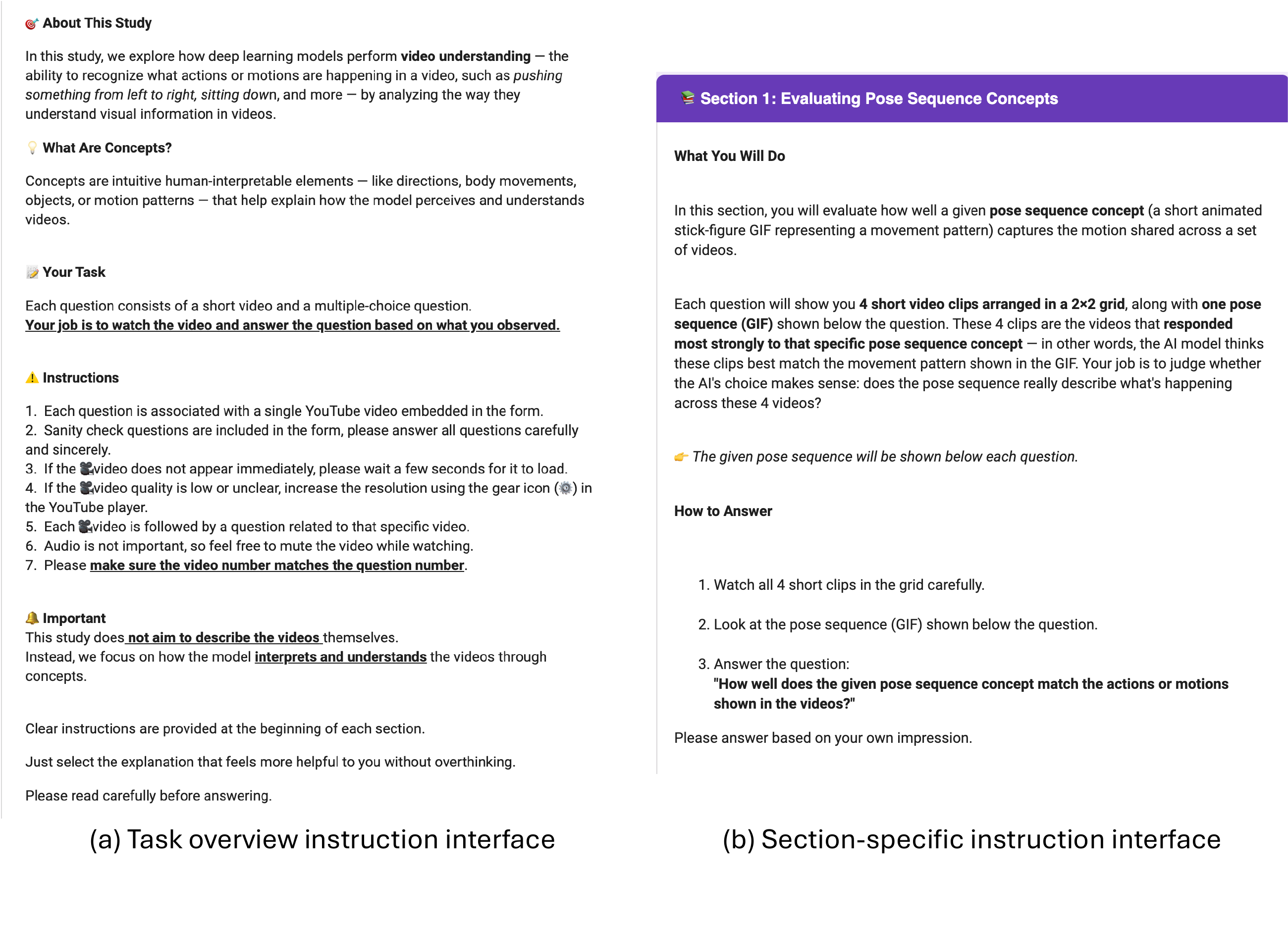}
    \vspace{-1em}
    \figcaption{Instruction interfaces shown to participants}{
    (a) We provide overview instruction at the beginning of the study.  
    (b) We provide section-specific instructions before each task to guide participants throughout the study.
    }
    \label{fig:userstudy_instruction}
\end{figure}

\begin{figure*}[t]
\centering
\includegraphics[width=1.0\textwidth]{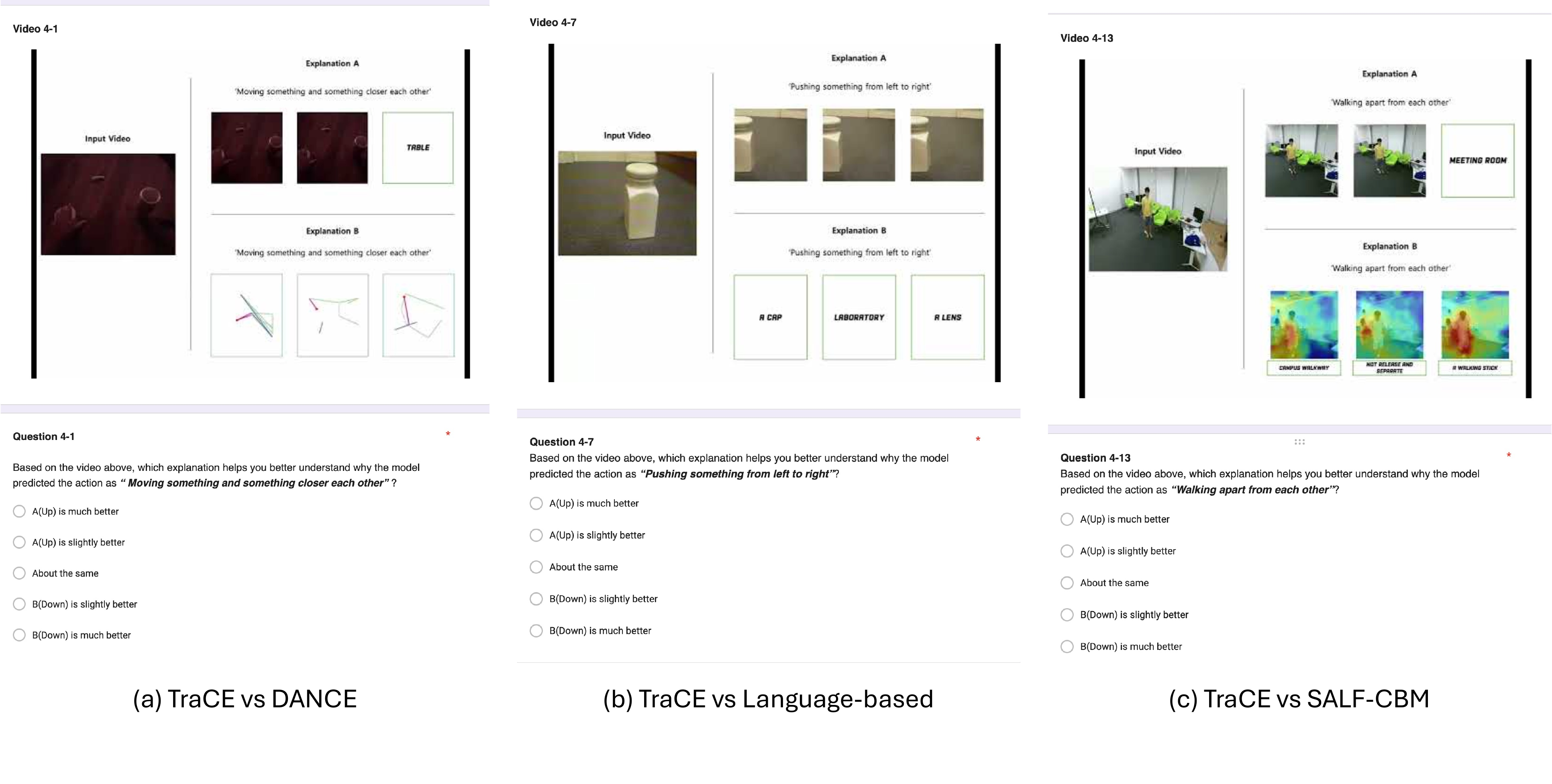}
    \figcaption{Example question and user interface used in Task 1}
    {Participants are asked to compare two explanations (A and B) for each sample or class.  
    (a) DANCE~\cite{dance}, (b) Language-based~\cite{gpt5}, and (c) SALF-CBM~\cite{salf-cbm}.  
    In (a) and (b), we show a single video sample with top-3 sample-level concept contributions.  
    In (c), SALF-CBM provides a heatmap-based spatial explanation alongside concept labels.}
    \vspace{\figcapmargin}
    \label{fig:task1_us}
\end{figure*}

\begin{figure*}[t]
\centering
\includegraphics[width=1.0\textwidth]{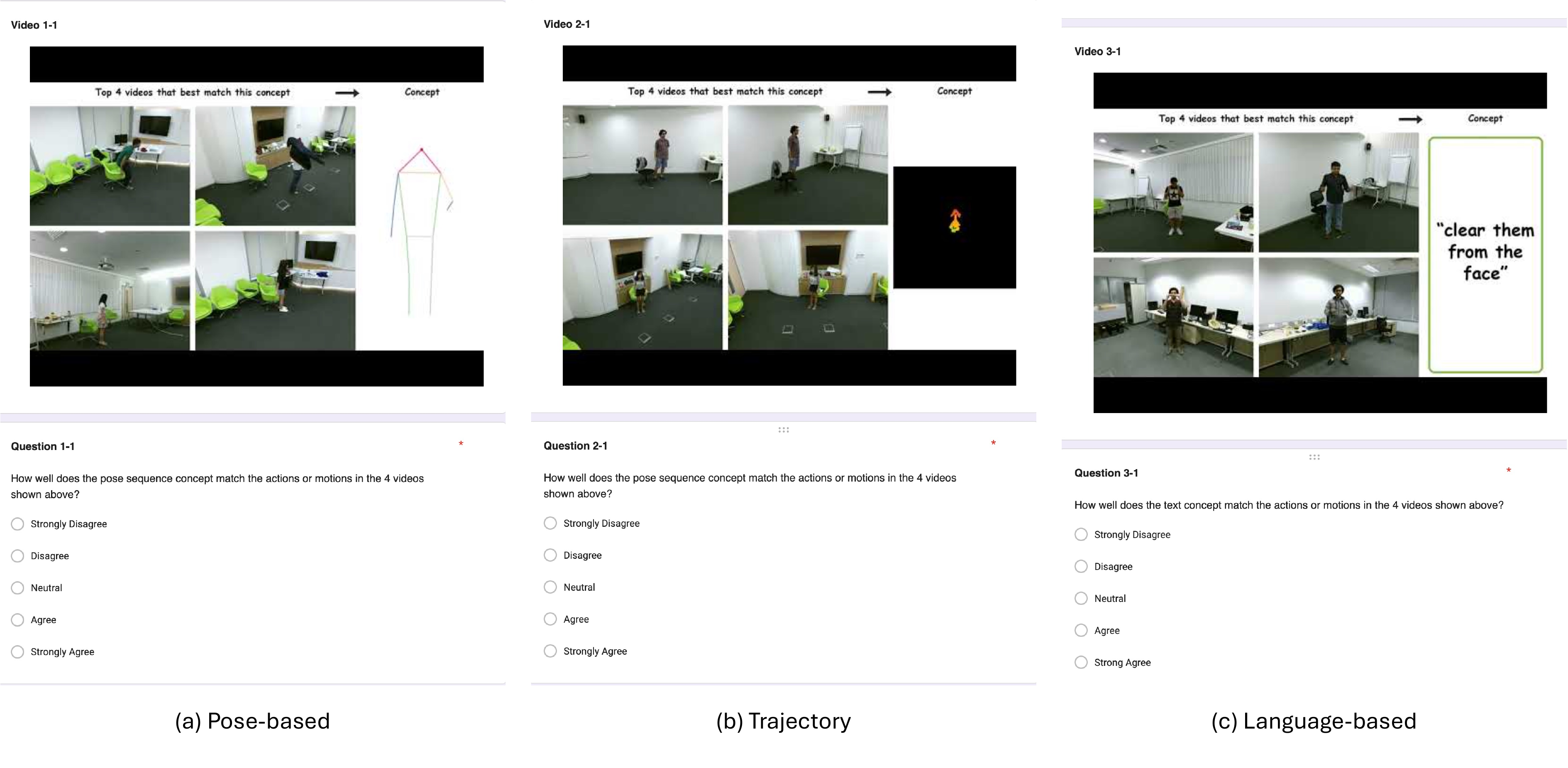}
    \figcaption{Example question and user interface used in Task 2}{We show how participants evaluate the interpretability of a single concept using several highly activated video clips.  
    Each interface corresponds to one of the three concept types.}
    \vspace{\figcapmargin}
    \label{fig:task2_us}
\end{figure*}
 
We provide clear instructions to participants prior to each task, as illustrated in \figref{userstudy_instruction}.
At the beginning of the study, participants receive an overview instruction page explaining that concepts are intuitive human-interpretable elements---such as body movements, objects, or motions---that help explain how the model perceives and understands the videos.
We emphasize that the study does not aim to describe the videos themselves, but to evaluate how well the model interprets and understands the videos through concepts.
Before each task, section-specific instructions describe the evaluation criteria and how to answer.
 
\paragraph{Task 1 interface. }
For each pairwise comparison question, we present a video with two explanations (Explanation A and Explanation B) side by side.
Each explanation shows the top-3 contributing concepts overlaid on the input video.
Participants are asked: ``\emph{Which explanation better helps you understand why the model predicted the action?}''
Responses are collected on a five-point scale: \emph{A is much better}, \emph{A is slightly better}, \emph{About the same}, \emph{B is slightly better}, and \emph{B is much better}.
We show an example of the Task~1 evaluation interface in \figref{task1_us}.
 
\paragraph{Task 2 interface. }
For each concept evaluation question, we present a single concept alongside its top-4 most strongly activating video clips.
Participants are asked: ``\emph{How well does the given concept match the actions or motions shown in the video above?}''
Responses are collected on a five-point Likert scale (1: Strongly Disagree to 5: Strongly Agree).
Each interface corresponds to one of the three concept types: trajectory concepts (\ours{}), pose sequence concepts (DANCE), or language-derived motion concepts (GPT-5.4).
We show an example of the Task~2 evaluation interface in \figref{task2_us}.

\subsection{Additional Results}
\label{appen:userstudy_stats}
\begin{table}[t]
\centering
\caption{\textbf{Mean Likert scores for Task~2 (concept quality).} We report the average score (1: Strongly Disagree to 5: Strongly Agree) over $300$ responses per condition. The \emph{original} condition shows each concept with its top-4 most strongly activating video clips; the \emph{random} condition shows a randomly selected concept with the same clips. $\Delta$ denotes the change from original to random (random $-$ original).}
\label{tab:userstudy_dist}
\small
\begin{tabular}{lccc}
\toprule
Method & Original & Random & $\Delta$ \\
\midrule
DANCE~\cite{dance}          & 2.52 & 2.55 & $0.03$ \\
\ours{}    & \textbf{4.22} & 2.67 & $-1.55$ \\
Language-based~\cite{gpt5}  & 2.27 & 1.86 & $-0.41$ \\
\bottomrule
\end{tabular}
\end{table}

To further validate the user study results in \figref{us1} (b) of the main paper, we report mean Likert scores under two conditions in \tabref{userstudy_dist}.
In the \emph{original} condition, participants evaluate a concept alongside its top-4 most strongly activating video clips.
In the \emph{random} condition, we keep the same highly activated video clips but replace the displayed concept with a randomly selected concept from the same method.
If a concept type is genuinely interpretable, the original condition should receive substantially higher ratings than the random condition, because the displayed concept should match the motion observed in the clips.
\ours{} achieves the highest mean score of $4.22$ under the original condition, indicating that participants strongly agree that trajectory concepts match the observed motion.
When the concept is replaced with a random one, the mean drops sharply to $2.67$ ($\Delta = -1.55$), confirming that the high ratings reflect genuine concept--motion alignment rather than a response bias.
In contrast, DANCE shows almost no difference between the original ($2.52$) and random ($2.55$) conditions ($\Delta = +0.03$).
We attribute this to the nature of the evaluation datasets: NTU RGB+D Single contains many simple, repetitive single-person motions (e.g., sitting down, standing up), so different pose sequence concepts tend to look similar to participants, making it difficult to distinguish the correct concept from a random one.
The language-based baseline receives the lowest original score ($2.27$), and its random condition score drops modestly ($\Delta = -0.41$), indicating that participants can sometimes distinguish correct language concepts from random ones, but the overall interpretability remains low.
Together, these results support the conclusion in the main paper that trajectory concepts are more interpretable and better aligned with underlying motion patterns than both pose-based and language-based alternatives.

\subsection{Pairwise User Study in the VQA Setting}
\label{appen:vqa_user_study}

The user study in \figref{us1} evaluates \ours{} explanations for action recognition.
We extend the evaluation to question-conditioned predictions through a blinded pairwise study with $50$ adult participants recruited through Prolific.
To isolate explanation quality from answer correctness, we include only examples for which both compared models produce the same correct answer.
Each item presents an identical video, question, answer candidates, and model answer together with two concept-based explanations.
Both explanations contain the same number of high-contribution concepts and use the same presentation format; the method identities are hidden.
Participants answer, ``Which explanation better helps you understand why the model chose its answer?''

Each participant evaluates five comparisons against the MM-CBM~\cite{mm-cbm} variant with language-derived motion concepts and five against the MM-CBM variant with pose-sequence concepts, yielding $250$ participant--item ratings per comparison.
The examples span the native TOMATO and STAR benchmarks and multiple-choice VQA reformulations of NTU RGB+D Mutual and UCF-101.

\begin{table}[H]
\centering
\caption{\textbf{Pairwise explanation preference in the VQA setting.}
Each comparison contains $250$ participant--item ratings from examples for which both models produce the same correct answer.
The five response columns are ordered from \ours{} being much better to the baseline being much better; the final column combines the first two responses.}
\label{tab:vqa_user_study}
\small
\setlength{\tabcolsep}{3pt}
\resizebox{\linewidth}{!}{%
\begin{tabular}{lcccccc}
\toprule
Comparison & Much better & Slightly better & About the same & Slightly worse & Much worse & Prefer \ours{} \\
\midrule
Language-derived motion concepts
& 123 (49.2\%) & 53 (21.2\%) & 25 (10.0\%) & 34 (13.6\%) & 15 (6.0\%) & \textbf{70.4\%} \\
Pose-sequence concepts
& 167 (66.8\%) & 47 (18.8\%) & 23 (9.2\%) & 10 (4.0\%) & 3 (1.2\%) & \textbf{85.6\%} \\
\bottomrule
\end{tabular}%
}
\end{table}

As shown in \tabref{vqa_user_study}, participants prefer \ours{} in $70.4\%$ of comparisons against explanations based on language-derived motion concepts and $85.6\%$ against explanations based on pose-sequence concepts.
Both models are correct on every evaluated item and display the same answer, so the preferences measure the perceived helpfulness of the explanations rather than differences in prediction correctness.
Because the study contains both native and multiple-choice reformulated examples, we describe it as a VQA-setting study; native VQA faithfulness is evaluated separately in \apref{native_vqa}.

\section{Additional Experimental Results}
\label{appen:add_results}
\label{appen:review_analyses}

For fair comparison, all experiments in this section use the same backbone, VideoMAE-B/16~\cite{videomae} for action recognition and InternVideo2$_{s2}$-1B~\cite{wang2024internvideo2} for video question answering unless otherwise noted. Unless specified otherwise, all ablation studies are conducted on Chiral SSv2~\cite{ssv2, chirality} dataset.

\subsection{Trajectory Concept Generation Ablation}
\label{appen:abl_trajectory}

\paragraph{Effect of number of concepts.}
We ablate the effect of the number of trajectory concepts $M_t$ by varying the FINCH partition level.
A higher partition level yields fewer, coarser clusters, while a lower level produces more, finer-grained clusters.
As shown in \tabref{concept_num}, the default configurations listed in \tabref{concept_config} achieve accuracy within $3$ percentage points of the finest partition across all datasets, while using substantially fewer concepts.
On NTU RGB+D Mutual, accuracy is nearly invariant to the number of concepts ($97.5$\% at both $681$ and $121$ concepts), reflecting the near-saturated nature of this benchmark.
On KTH, the default partition ($M_t = 111$) slightly outperforms the finest partition ($M_t = 589$, $92.0$\% vs.\ $91.0$\%), suggesting that an excessive number of fine-grained concepts can introduce noise for datasets with simple, well-separated motion patterns.
In contrast, Chiral SSv2 shows the largest sensitivity to concept count ($86.8$\% $\rightarrow$ $80.3$\% from $870$ to $29$ concepts), which is expected because its paired directional classes require fine-grained trajectory discrimination.

\begin{table}[H]
\centering
\caption{\textbf{Trajectory concept generation ablations.} (a) is reported on all action recognition datasets; (b)--(f) are reported on Chiral SSv2. Default settings are \ours{}'s configuration in \tabref{concept_config}.}
\label{tab:traj_ablation}
\begin{subtable}{\linewidth}
\centering
\caption{Number of trajectory concepts.}
\label{tab:concept_num}
\small
\begin{tabular}{cc cc cc cc}
\toprule
\multicolumn{2}{c}{Chiral SSv2}
& \multicolumn{2}{c}{NTU RGB+D Mutual}
& \multicolumn{2}{c}{UCF-101}
& \multicolumn{2}{c}{KTH} \\
\cmidrule(lr){1-2}
\cmidrule(lr){3-4}
\cmidrule(lr){5-6}
\cmidrule(lr){7-8}
\# Concepts & Acc.
& \# Concepts & Acc.
& \# Concepts & Acc.
& \# Concepts & Acc. \\
\midrule
870  & 86.8 & 681 & 97.5 & 4124 & 89.5 & 589 & 91.0 \\
154  & 84.2 & 121 & 97.5 & 759  & 88.6 & 111 & 92.0 \\
29   & 80.3 & 25  & 96.9 & 134  & 88.4 & 21  & 90.4 \\
\bottomrule
\end{tabular}
\end{subtable}

\vspace{0.6em}
\small
\setlength{\tabcolsep}{3pt}
\begin{subtable}[t]{0.13\linewidth}
\centering
\caption{$N_s$.}
\label{tab:segment}
\begin{tabular}{lc}
\toprule
$N_s$ & Acc. \\
\midrule
1 & 83.2 \\
3 & 84.2 \\
5 & 82.6 \\
\bottomrule
\end{tabular}
\end{subtable}
\hfill
\begin{subtable}[t]{0.13\linewidth}
\centering
\caption{$L$.}
\label{tab:length}
\begin{tabular}{lc}
\toprule
$L$ & Acc. \\
\midrule
8  & 86.9 \\
16 & 84.2 \\
32 & 87.3 \\
\bottomrule
\end{tabular}
\end{subtable}
\hfill
\begin{subtable}[t]{0.21\linewidth}
\centering
\caption{Clustering.}
\label{tab:clustering}
\begin{tabular}{lc}
\toprule
Method & Acc. \\
\midrule
FINCH~\cite{finch} & 84.2 \\
K-Means~\cite{k-means} & 84.6 \\
\bottomrule
\end{tabular}
\end{subtable}
\hfill
\begin{subtable}[t]{0.23\linewidth}
\centering
\caption{Labeling.}
\label{tab:soft_vs_hard}
\begin{tabular}{lc}
\toprule
Method & Acc. \\
\midrule
Binary & 82.4 \\
Soft & 84.2 \\
Max normalized & 83.3 \\
\bottomrule
\end{tabular}
\end{subtable}
\hfill
\begin{subtable}[t]{0.21\linewidth}
\centering
\caption{Foreground mask.}
\label{tab:foreground_gating}
\begin{tabular}{lc}
\toprule
Masking & Acc. \\
\midrule
On (\ours{}) & 84.2 \\
Off & 80.3 \\
\bottomrule
\end{tabular}
\end{subtable}
\end{table}

\paragraph{Effect of the number of segments.}
As shown in \tabref{segment}, accuracy is robust to the number of temporal segments $N_s$, with $N_s = 3$ achieving the best result ($84.2$\%).
Using a single segment ($N_s = 1$) slightly reduces accuracy ($83.2$\%) because the entire video is summarized by one clip, losing temporal coverage.
Increasing to \(N_s=5\) also decreases accuracy to 82.6\%.
We set $N_s$ per dataset by dividing the average number of frames in the training set by the trajectory length $L$, which provides a natural balance between temporal coverage and clip quality.

\paragraph{Effect of trajectory length $L$.}
\begin{figure*}[t]
\includegraphics[width=\textwidth]{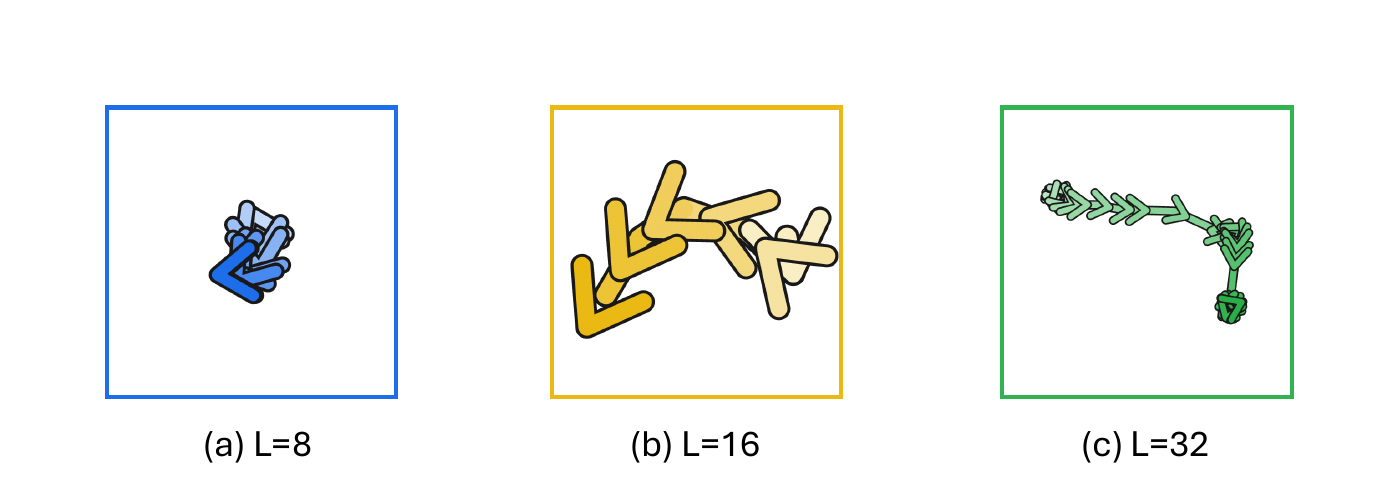}
\caption{
\textbf{Trajectory concepts discovered with different trajectory lengths.}
Short trajectories (\(L = 8\)) capture only local and fragmented motion patterns with limited temporal structure.
Long trajectories (\(L = 32\)) often encode near-complete action dynamics, reducing concept reusability across classes.
In contrast, the default trajectory length (\(L = 16\)) captures representative motion fragments that remain both interpretable and reusable across different videos.
These visualizations illustrate the trade-off between local motion specificity and action-level over-specialization when choosing the trajectory length.
}
\label{fig:length}
\end{figure*}

As shown in \tabref{length}, longer trajectories ($L = 32$) achieve the highest accuracy ($87.3$\%), while the default length ($L = 16$) achieves $84.2$\% and shorter trajectories ($L = 8$) achieve $86.9$\%.
However, accuracy alone does not reflect concept interpretability.
We visualize trajectory concepts discovered at each length in \figref{length}.
Short trajectories ($L = 8$) capture only brief, fragmented motion that lacks sufficient temporal context to convey a meaningful motion pattern.
Long trajectories ($L = 32$) tend to encode the entire action class rather than reusable motion primitives shared across classes---analogous to classifying an image by the object itself rather than by its distinguishing attributes.
We therefore adopt $L = 16$ as the default, which captures representative motion fragments that are both discriminative and generalizable as an interpretable trajectory vocabulary.

\paragraph{Effect of clustering method.}
We compare FINCH~\cite{finch} with K-Means~\cite{k-means} as the trajectory clustering algorithm.
As shown in \tabref{clustering}, the two methods achieve comparable accuracy ($84.2$\% vs.\ $84.6$\%), confirming that trajectory concept quality is largely agnostic to the choice of clustering algorithm.
We adopt FINCH as the default because it automatically determines the number of clusters without requiring a predefined $k$, making the pipeline easier to apply across datasets without per-dataset hyperparameter tuning.

\paragraph{Soft vs.\ hard concept labels.}
We compare three strategies for converting trajectory cluster assignments into video-level concept labels: (i)~\emph{Binary}, which assigns $1$ if any trajectory in the video belongs to the cluster and $0$ otherwise; (ii)~\emph{Soft} (default), which uses the normalized frequency $\mathbf{c}_i^t = \mathbf{h}_i / \|\mathbf{h}_i\|_1$; and (iii)~\emph{Max normalized}, which divides each count by the maximum count across all concepts for the video.
As shown in \tabref{soft_vs_hard}, the soft labeling strategy achieves the best accuracy ($84.2$\%), outperforming both binary ($82.4$\%) and max normalized ($83.3$\%) alternatives.
Binary labels discard frequency information and treat rare and dominant motion patterns equally, which reduces the model's ability to distinguish fine-grained directional cues.

\paragraph{Effect of foreground masking.}
The trajectory filter in \apref{trajectory_concept} retains a trajectory only when its seed pixel lies inside the estimated foreground saliency mask.
On Chiral SSv2, foreground masking removes approximately $81\%$ of the raw trajectories.
As shown in \tabref{foreground_gating}, foreground masking improves validation accuracy from $80.3\%$ to $84.2\%$ when we keep the remaining pipeline unchanged, supporting its use to suppress background-seeded trajectories.

\subsection{Analysis}
\label{appen:app_analysis}

\paragraph{Effect of Video-LLM.}
\begin{table}[ht]
    \centering
    \caption{\textbf{Effect of different object and scene concept discovery method.} }

    \vspace{0.5em}
    \resizebox{0.55\linewidth}{!}{
    \begin{tabular}{ l r r r}
        \toprule
        Method &&  \# Concepts & Acc.\\
        \midrule
        GPT-5.4~\cite{gpt5} && 152 & 61.5 \\
        LLaVA-NeXT-Video~\cite{llavanext-video}  && 393 & 71.8 \\
        Video-LLaVA~\cite{videollava}  && 1,026 & 77.7\\
        VideoLLaMA2~\cite{videollama2} && 1,478 & 77.3\\
        Qwen3-VL~\cite{qwen3vl} && 1,683 & 74.6 \\

        \bottomrule
    \end{tabular}
    }
    \label{tab:text_concept}
\end{table}

\begin{figure*}[t]
\includegraphics[width=\textwidth]{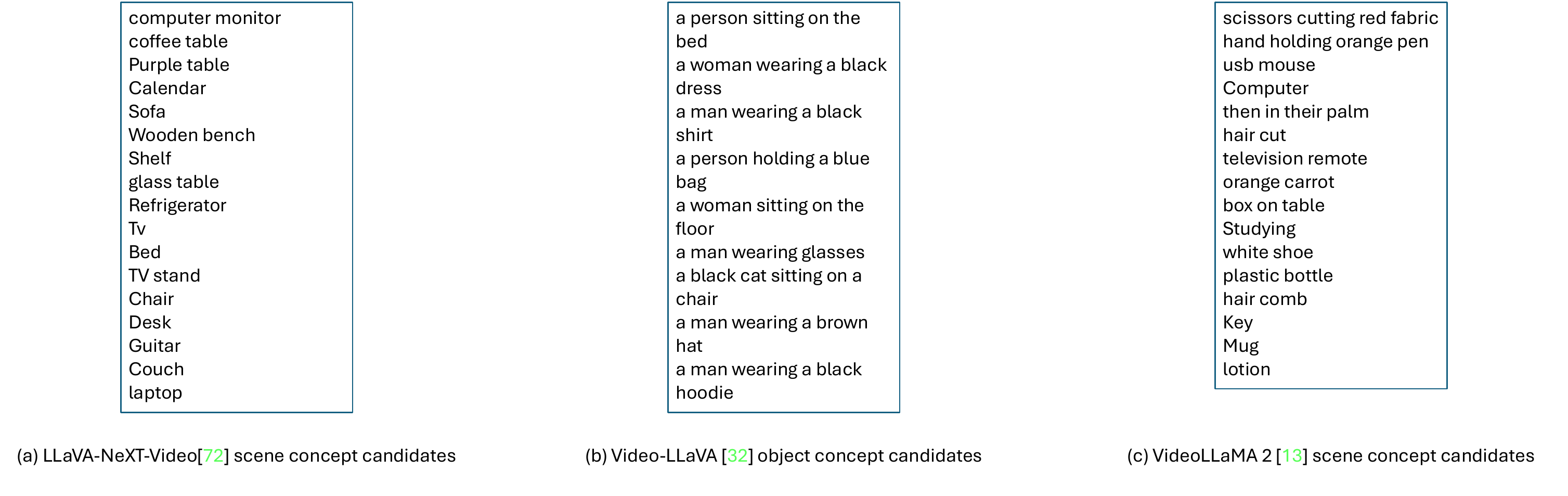}
\caption{
\textbf{Qualitative comparison of video-language models for text concept generation.}
We visualize representative object and scene concept candidates generated by different video-language models.
(a) LLaVA-NeXT-Video~\cite{llavanext-video} produces relatively clean scene-centric concepts.
(b) Video-LLaVA~\cite{videollava} frequently mixes appearance descriptions and human-centric phrases into object concepts.
(c) VideoLLaMA2~\cite{videollama2} sometimes generates action phrases or interaction descriptions instead of static scene concepts.
These examples highlight that higher downstream accuracy does not necessarily imply cleaner object--scene concept separation.
}
\label{fig:text_concept_quality}
\end{figure*}

We compare five language models for generating object and scene concept candidates: GPT-5.4~\cite{gpt5}, LLaVA-NeXT-Video~\cite{llavanext-video}, Video-LLaVA~\cite{videollava}, VideoLLaMA2~\cite{videollama2}, and Qwen3-VL~\cite{qwen3vl}.
As shown in \tabref{text_concept}, Video-LLaVA and VideoLLaMA2 achieve the highest accuracy ($77.7$\% and $77.3$\%), while GPT-5.4 produces the fewest concepts ($152$) and the lowest accuracy ($61.5$\%).
Qwen3-VL achieves competitive accuracy ($74.6$\%) despite generating the most candidates ($1{,}683$).
However, accuracy alone does not capture concept quality.
We qualitatively observe that many models produce concepts that violate the typed separation between object and scene groups: for example, scene-like terms appear in the object list, or action verbs are included among object candidates.
We show representative examples of such violations in \figref{text_concept_quality}.
We adopt Qwen3-VL as the default because it produces the cleanest separation between object and scene concepts when guided by our structured prompts (\figref{prompt}), even though other models may achieve slightly higher accuracy with noisier concept vocabularies.

\paragraph{Vision-language model ablation.}
\label{appen:abl_vlm}
\begin{table}[t]
    \centering
    \caption{\textbf{Evaluation of VLM encoder for object and scene concept labeling} }

    \resizebox{0.6\linewidth}{!}{
        \begin{tabular}{l l c }
        \toprule
         Video encoder & VLM encoder  & Acc. \\
        \midrule
         \multirow{2}{*}{VideoMAE ~\cite{videomae}} & + InternVid~\cite{internvid}  & 83.4 \\
         & + InternVideo2~\cite{wang2024internvideo2} & 84.2 \\
        \bottomrule
        \end{tabular}
        }
 \label{tab:VLM_encoder}

\end{table}

We ablate the choice of the vision--language encoder used to compute the object and scene pseudo-labels (\apref{object_and_scene_concept}), while keeping the same concept candidates and backbone.
As shown in \tabref{VLM_encoder}, replacing InternVid~\cite{internvid} with InternVideo2~\cite{wang2024internvideo2} as the labeling encoder improves accuracy from $83.4$\% to $84.2$\%.
We attribute this improvement to the stronger video--text alignment provided by InternVideo2's Stage-2 contrastive pretraining, which produces more accurate soft pseudo-labels for each concept.

\paragraph{Disentanglement analysis.}
\label{appen:abl_disentangle}
We analyze whether the three concept groups (trajectory, object, scene) use disentangled subspaces of the backbone feature $\mathbf{x}_i$.
For this additional analysis, we ablate two orthogonality mechanisms: (1) \emph{SVD null-space projection}, which removes previously used subspace dimensions from the backbone feature before training the next concept group, and (2) \emph{$\mathbf{W}_c$ row-orthogonality loss} with weight $\lambda$, which penalizes cosine similarity between weight rows of different groups.
We measure disentanglement via top-$K$ backbone-dimension overlap: for each group $g$, we compute a per-dimension importance score $\text{imp}_g[j] = \text{mean}_n |\mathbf{x}_n[j]| \cdot \text{mean}_c |\mathbf{W}_c^g[c, j]|$ over validation samples, select the top-$K$ dimensions, and count shared indices across group pairs.
The random-chance baseline for $K = 50$ is $K^2 / d = 3.26$ shared dimensions.
\begin{table}[t]
\centering
\caption{\textbf{Component ablation for concept group disentanglement.} We report validation accuracy (\%) and top-$50$ backbone-dimension overlap between group pairs. Random-chance overlap is $3.26$. The object--scene pair (object $\cap$ scene) is the structurally hardest target because both groups are text-based visual concepts.}
\label{tab:disent_component}
\small
\setlength{\tabcolsep}{4pt}
\begin{tabular}{lcccc}
\toprule
Variant & Acc.\ (\%) & traj $\cap$ obj & traj $\cap$ scene & obj $\cap$ scene \\
\midrule
Baseline                     & 84.20 & 12 & 10 & 34 \\
SVD only ($\alpha{=}1.0$)    & 84.55 & 10 & 19 & 19 \\
Ortho only ($\lambda{=}1.0$) & 86.92 & 8  & 9  & 21 \\
SVD + Ortho ($\lambda{=}1.0$)& 86.78 & 8  & 9  & 22 \\
SVD + Ortho ($\lambda{=}2.0$)& \textbf{87.83} & 8  & 8  & 26 \\
\bottomrule
\end{tabular}
\end{table}

\subparagraph{Group ordering sensitivity.}
Since concept groups are trained sequentially, we fix SVD + Ortho ($\lambda = 1.0$) and train all six permutations of \{trajectory (T), object (O), scene (S)\}.
As shown in \tabref{disent_order}, the ordering induces a $1.2$\%p accuracy spread and a $1.7\times$ variation in maximum pairwise top-$50$ overlap.
\emph{Trajectory-first} orders (TSO, TOS) achieve the highest average accuracy ($86.85$\%), because the trajectory branch uses a structurally distinct BCE head and naturally occupies a separate subspace, imposing no representational cost on subsequent groups.
\emph{Object-first} orders (OTS, OST) yield the lowest accuracy ($85.94$\% average), as the object group has the largest concept set ($404$ concepts) and consumes a large subspace first.
SOT achieves the lowest maximum overlap ($13$) and lowest sum overlap ($29$), while TSO achieves the best accuracy ($86.92$\%) with a moderate maximum overlap ($20$).
We use TOS for the component ablation in \tabref{disent_component}.

\subparagraph{Component ablation.}
We compare four variants against the baseline (no orthogonality) in \tabref{disent_component}.
The $\mathbf{W}_c$ orthogonality loss is the dominant driver of both accuracy and disentanglement.
SVD alone contributes only $+0.35$\%p in accuracy and becomes redundant at inference when the orthogonality loss is active: for all ortho-containing variants, removing the SVD projection at inference causes no accuracy change ($\Delta = 0$).
Combining SVD with orthogonality at $\lambda = 2.0$ achieves the best accuracy ($87.83$\%).
All variants reduce object--scene overlap relative to the baseline ($34 \to 19$--$26$), confirming that orthogonality mechanisms successfully encourage the two text-based groups to use more distinct backbone dimensions.
Increasing $\lambda$ from $1.0$ to $2.0$ improves accuracy ($+1.0$\%p) but slightly increases top-$50$ overlap ($22 \to 26$), because row-orthogonality can be satisfied by using the same dimensions with opposite sign or scale.
\begin{table}[t]
\centering
\caption{\textbf{Effect of concept group training order on accuracy and disentanglement.} We report validation accuracy (\%) and top-$50$ backbone-dimension overlap between group pairs for all six permutations of \{trajectory (T), object (O), scene (S)\}. SVD + Ortho ($\lambda = 1.0$) is fixed.}
\label{tab:disent_order}
\small
\setlength{\tabcolsep}{4pt}
\begin{tabular}{lccccc}
\toprule
Order & Acc.\ (\%) & traj $\cap$ obj & traj $\cap$ scene & obj $\cap$ scene & sum \\
\midrule
TSO & \textbf{86.92} & 10 & 9  & 20 & 39 \\
TOS & 86.78 & 8  & 9  & 22 & 39 \\
STO & 86.64 & 12 & 16 & 12 & 40 \\
SOT & 86.50 & 13 & 7  & \textbf{9}  & \textbf{29} \\
OTS & 86.15 & 14 & 8  & 12 & 34 \\
OST & 85.73 & 12 & 13 & 11 & 36 \\
\bottomrule
\end{tabular}
\end{table}

\paragraph{Concept combination.}
\label{appen:abl_composition}

\begin{table}[t]
    \centering
    \caption{\textbf{Effects of individual concept types on \ours{} performance.}}
    \vspace{0.2em}
    \resizebox{0.45\linewidth}{!}{ %
        \begin{tabular}{ccccc}
        \toprule
         \multicolumn{3}{c}{Concept} & \multirow{2}{*}{Acc.}  \\
         \cmidrule(lr){1-3} %
        Trajectory & Object & Scene &  \\
        \midrule
         \ding{51} &  \ding{55} & \ding{55} &  80.7 \\
         \ding{51}& \ding{51} & \ding{55} &  82.4 \\
        \ding{51} & \ding{55} & \ding{51} & 81.6\\
        \ding{55} & \ding{51} & \ding{51} &   74.6 \\
        \ding{51} & \ding{51} & \ding{51} & 84.2\\
        \bottomrule
        \end{tabular}
    }
    \label{tab:concept_combination}

\vspace{-1.5em}
\end{table}

We evaluate the contribution of each concept type by training \ours{} with different subsets of the three concept groups.
As shown in \tabref{concept_combination}, using all three groups (trajectory + object + scene) achieves the best accuracy ($84.2$\%).
Trajectory concepts alone already capture the majority of the performance ($80.7$\%), and adding object or scene concepts yields consistent gains ($+1.7$\%p and $+0.9$\%p, respectively).
In contrast, removing trajectory concepts entirely (object + scene only) causes a substantial drop to $74.6$\%, a $-9.6$\%p gap relative to the full model.
This confirms that trajectory concepts are the most informative group, while object and scene concepts serve as useful but secondary complements.

\paragraph{Concept sparsity.}
\label{appen:abl_leakage}
\begin{table}[t]
\centering
\caption{\textbf{Accuracy under controlled Number of Effective Concepts (NEC)~\cite{vlgcbm}.} NEC limits the average number of non-zero classifier weights per class. Lower NEC yields sparser, more interpretable predictions at the cost of accuracy. Sparsity denotes the fraction of non-zero entries in $\mathbf{W}_{\text{AR}}$.}
\label{tab:info_leakage}
\small
\begin{tabular}{ccc}
\toprule
NEC & Sparsity (\%) & Acc.\ (\%) \\
\midrule
5  &  0.63 & 67.7 \\
10 &  1.21 & 77.2 \\
15 &  1.79 & 81.3 \\
20 &  2.41 & 82.9 \\
25 &  3.03 & 83.8 \\
30 &  3.57 & 84.0 \\
\bottomrule
\end{tabular}
\end{table}

We evaluate whether \ours{} maintains predictive performance with a compact set of concepts using the Number of Effective Concepts (NEC) metric proposed by VLG-CBM~\cite{vlgcbm}.
We vary the regularization strength $\lambda$ of the GLM-SAGA solver to control the average number of non-zero weights per class in $\mathbf{W}_{\text{AR}}$.
As shown in \tabref{info_leakage}, at NEC $= 25$, accuracy reaches $83.8$\%, within $0.4$ percentage points of the default model ($84.2$\%).
Even at NEC $= 5$, where each class uses only five concepts on average, accuracy remains at $67.7$\%.

\subsection{Explanation Analysis}
\label{appen:explanation_analysis}

\paragraph{Motion perturbation analysis.}

\begin{table}[t]
\centering
\caption{
\textbf{Prediction consistency under perturbation on Chiral SSv2.}
We report the percentage of samples whose predicted motion group follows the expected transformation after perturbation.
Expected transformations are:
LEFT $\leftrightarrow$ RIGHT and UP $\leftrightarrow$ DOWN under temporal reverse,
and LEFT $\to$ DOWN, RIGHT $\to$ UP, UP $\to$ LEFT, DOWN $\to$ RIGHT under 90$^\circ$ counter-clockwise rotation.
}
\label{tab:prediction_consistency}
\begin{tabular}{lcc}
\toprule
Original group
& Reverse consistency (\%)
& Rotation consistency (\%) \\
\midrule
Left  & 90.4 & 83.3 \\
Right & 84.0 & 90.8 \\
Up    & 55.8 & 44.2 \\
Down  & 94.0 & 45.2 \\
\midrule
Overall & 79.2 & 63.1 \\
\bottomrule
\end{tabular}
\end{table}

We complement the perturbation analysis in \secref{edit} with direction-specific results on the same directional-class subset of Chiral SSv2.
We evaluate classes specifying left, right, up, or down motion under temporal reversal and $90^\circ$ counter-clockwise rotation.
As shown in \figref{sanity}, \ours{} adapts its top trajectory concepts to the induced motion transformation.
For the original input in \figref{sanity} (a), the top concepts align with the left-to-right motion and support the prediction.
When the video is temporally reversed in \figref{sanity} (b), the top concepts switch to the opposite direction and the prediction changes accordingly.
Under a \(90^\circ\) rotation in \figref{sanity} (c), the activated concepts rotate consistently with the transformed motion direction.
For each test sample, we apply the perturbation and check whether the predicted motion group follows the expected directional transformation: temporal reversal should swap LEFT $\leftrightarrow$ RIGHT and UP $\leftrightarrow$ DOWN, while $90^\circ$ rotation should map LEFT $\to$ DOWN, RIGHT $\to$ UP, UP $\to$ LEFT, and DOWN $\to$ RIGHT.
As shown in \tabref{prediction_consistency}, \ours{} achieves an overall reverse consistency of 79.2\% and a rotation consistency of 63.1\%.
Reverse consistency is higher for LEFT (90.4\%), RIGHT (84.0\%), and DOWN (94.0\%) than for UP (55.8\%).
Rotation consistency is also higher for LEFT (83.3\%) and RIGHT (90.8\%) than for UP (44.2\%) and DOWN (45.2\%).
The prediction-level results support motion sensitivity and complement the concept-level directional analysis in \apref{directional_semantics}.

\begin{table}[t]
\centering
\caption{
\textbf{Faithfulness of trajectory concepts via concept removal. }
We report Flip\%, the percentage of samples whose prediction changes after deactivating trajectory concepts.
We compare removing the top-K contributing concepts with removing K randomly selected concepts.
Higher Top-K removal and lower Random-K removal indicate that the estimated concept importance aligns with the model's actual decision process.
}
\label{tab:faithfulness}
\begin{tabular}{l cc cc}
\toprule
& \multicolumn{2}{c}{K=5}
& \multicolumn{2}{c}{K=10} \\
\cmidrule(lr){2-3}
\cmidrule(lr){4-5}
Dataset
& Top-K $\uparrow$
& Random-K $\downarrow$
& Top-K $\uparrow$
& Random-K $\downarrow$ \\
\midrule
Chiral SSv2         & 48.7 & 3.7 & 70.1 & 5.4 \\
KTH                 & 35.1 & 2.1 & 46.0 & 2.8 \\
UCF-101              & 15.9 & 0.2 & 18.4 & 0.3 \\
NTU RGB+D Mutual    & 7.0  & 0.4 & 13.8 & 0.3 \\
\midrule
Average             & 26.7 & 1.6 & 37.1 & 2.2 \\
\bottomrule
\end{tabular}
\end{table}

\paragraph{Faithfulness analysis.}
We extend the faithfulness evaluation in \secref{edit} of the main paper by reporting per-dataset concept removal results in \tabref{faithfulness}.
For each sample, we deactivate either the top-$K$ trajectory concepts ranked by contribution (Top-$K$) or $K$ randomly selected trajectory concepts (Random-$K$), and measure Flip\%, the percentage of samples whose predicted class changes.
A faithful model should show high Flip\% under Top-$K$ removal and low Flip\% under Random-$K$ removal, indicating that the contribution ranking aligns with the model's actual decision process.
Across all datasets, Top-$K$ removal causes substantially more prediction flips than Random-$K$ removal at both $K = 5$ and $K = 10$.
Chiral SSv2 shows the largest Top-$K$ effect ($48.7$\% at $K{=}5$, $70.1$\% at $K{=}10$), consistent with the strong dependence of directional motion classes on a small number of discriminative trajectory concepts.
NTU RGB+D Mutual shows the smallest effect ($7.0$\% at $K{=}5$), reflecting the near-saturated nature of this benchmark where multiple trajectory concepts carry redundant evidence.
UCF-101 shows a moderate Top-$K$ effect ($15.9$\% at $K{=}5$), consistent with the observation that many UCF-101 classes can be supported by static scene or object cues alongside trajectory evidence.
In all cases, Random-$K$ removal causes minimal prediction changes ($\leq 5.4$\%), confirming that random trajectory concepts do not carry decision-critical information.

\subsection{Sample-Level Motion-Source Grounding}
\label{appen:source_grounding}

The qualitative examples in \figref{contrib} show that trajectory concepts explain interactions and object dynamics, while \tabref{generalizability} shows that the trajectory-only variant retains predictive motion cues under object category shift.
We quantify whether high-contribution trajectory concepts are spatially grounded in the hands and manipulated objects in each video.

We use the same object-disjoint Chiral SSv2 split constructed from Something-Else annotations~\cite{somethingelse}.
As described in \apref{concept_discovery}, we discover trajectory concepts exclusively from filtered training trajectories and retain the medoid trajectory of each training cluster.
For each validation video, we assign every cached trajectory to its nearest fixed medoid in the same flattened displacement space used for clustering.
The cached validation assignments are used to compute the validation loss and select the checkpoint, but validation trajectories do not participate in vocabulary construction or gradient updates.
We rank trajectory concepts by their contribution to the predicted class, select the trajectories assigned to the Top-$10$, Random-$10$, or Bottom-$10$ concepts, and recover their seed pixels in the input video.
For Random-$10$, we sample one set of ten concepts uniformly without replacement for each validation video and aggregate coverage over the validation set.
Because every assigned trajectory retains its seed pixel, this analysis directly recovers the spatial support of each selected concept without Grad-CAM.

Something-Else provides frame-level bounding boxes for hands and manipulated objects.
A selected trajectory covers a box only when its seed pixel lies inside the box in the first frame of the corresponding $L$-frame clip.
\emph{Hand} and \emph{Object} measure the percentages of annotated hand and object boxes covered by at least one selected trajectory.
\emph{Both} measures the percentage of videos in which the selected trajectories cover at least one annotated hand and one manipulated object.
We evaluate DANCE~\cite{dance} analogously using the spatial support of its selected pose-sequence concepts.

\begin{table}[H]
\centering
\caption{\textbf{Sample-level motion-source grounding on object-disjoint Chiral SSv2.}
Top and Bottom select $K=10$ concepts according to their contribution to the predicted class; Random samples one set of $K=10$ concepts uniformly without replacement for each video.
\emph{Hand} and \emph{Object} report the percentages of annotated boxes covered by the seed pixel of a selected trajectory in the first frame of its $L$-frame clip; \emph{Both} reports the percentage of videos covering at least one hand and one manipulated object.}
\label{tab:source_grounding}
\small
\setlength{\tabcolsep}{4.5pt}
\begin{tabular}{lcccccc}
\toprule
& \multicolumn{3}{c}{DANCE~\cite{dance}}
& \multicolumn{3}{c}{\ours{}} \\
\cmidrule(lr){2-4}\cmidrule(lr){5-7}
Selection & Hand & Object & Both & Hand & Object & Both \\
\midrule
Bottom-$10$ & 0.0 & 0.0 & 0.0 & 22.4 & 21.6 & 19.0 \\
Random-$10$ & 4.1 & 0.7 & 0.5 & 29.1 & 28.5 & 25.1 \\
Top-$10$    & 42.3 & 7.6 & 4.7 & \textbf{64.5} & \textbf{63.5} & \textbf{58.1} \\
\bottomrule
\end{tabular}
\end{table}

As shown in \tabref{source_grounding}, source coverage increases with concept contribution for \ours{}.
The Top-$10$ trajectory concepts cover 64.5\% of annotated hand boxes and 63.5\% of manipulated-object boxes, with both sources covered in 58.1\% of videos.
The joint coverage decreases to 25.1\% for Random-$10$ concepts and 19.0\% for Bottom-$10$ concepts.
In contrast, DANCE's Top-$10$ pose-sequence concepts cover 42.3\% of annotated hand boxes but only 7.6\% of manipulated-object boxes and both sources in 4.7\% of videos.
These results do not imply that each trajectory concept has a fixed entity label.
Rather, they show that high-contribution trajectory concepts can be grounded at the sample level to motion from both people and manipulated objects, supporting their role as a source-flexible motion vocabulary.

\subsection{Directional Semantic Validation}
\label{appen:directional_semantics}

As shown in \tabref{prediction_consistency_avg}, \ours{}'s predictions follow the expected directional transformations under temporal reversal and rotation.
Prediction consistency alone, however, does not show whether the selected trajectory concepts carry the corresponding motion semantics.
Because trajectory concepts are supervised by a soft distribution over multiple training-derived clusters, single-label precision or recall would discard the target structure.
We instead evaluate signed motion direction on the Chiral SSv2 direction subset ($n=572$)~\cite{ssv2,chirality}, whose actions specify \emph{left}, \emph{right}, \emph{up}, or \emph{down} motion.
We assign each trajectory concept a fixed direction label from the net displacement of its medoid trajectory before evaluating the predicted concept activations.
Across held-out videos, the highest-activated trajectory concept matches the ground-truth direction in 67.5\% of cases, compared with the 25\% four-way random reference.

For each original, temporally reversed, and $90^\circ$-rotated video, we measure whether the top-$5$ activated trajectory concepts follow the expected direction.
Let $C_{\mathrm{orig}}^5$ and $C_{\mathrm{pert}}^5$ denote the sets of top-$5$ concept indices for the original and perturbed videos.
We measure concept replacement as
$1-\lvert C_{\mathrm{orig}}^5 \cap C_{\mathrm{pert}}^5\rvert/5$.
For each transformed video, prediction correctness is determined using the expected transformed class.

\begin{table}[H]
\centering
\caption{\textbf{Directional semantics under controlled perturbations.}
Direction agreement is the fraction of top-$5$ activated trajectory concepts whose fixed direction labels, derived from their medoid trajectories, match the expected direction.
For the original and perturbed top-$5$ concept-index sets $C_{\mathrm{orig}}^5$ and $C_{\mathrm{pert}}^5$, replacement is $1-\lvert C_{\mathrm{orig}}^5 \cap C_{\mathrm{pert}}^5\rvert/5$.
Prediction correctness after perturbation is evaluated against the expected transformed class; the final row reports random-reference values.}
\label{tab:directional_semantics}
\small
\setlength{\tabcolsep}{5pt}
\begin{tabular}{llcc}
\toprule
Prediction & Input & Direction agreement (\%) & Replacement (\%) \\
\midrule
Correct   & Original             & 76.3 & 0.0 \\
Correct   & Temporal reversal    & \textbf{80.1} & \textbf{96.3} \\
Correct   & $90^\circ$ rotation & \textbf{78.4} & \textbf{96.1} \\
\midrule
Incorrect & Original             & 45.4 & 0.0 \\
Incorrect & Temporal reversal    & 32.8 & 81.0 \\
Incorrect & $90^\circ$ rotation & 35.1 & 84.6 \\
\midrule
Random reference & -- & 25.0 & 96.8 \\
\bottomrule
\end{tabular}
\end{table}

As shown in \tabref{directional_semantics}, correctly predicted transformed videos replace 96.3\% of the original top-$5$ concept identities under temporal reversal and 96.1\% under rotation.
The selected concepts achieve 80.1\% direction agreement after reversal and 78.4\% after rotation.
Incorrect transformed predictions provide a control: they also exhibit substantial concept replacement, but their direction agreement decreases to 32.8\% under reversal and 35.1\% under rotation.
Concept replacement alone therefore does not establish semantic consistency.
Correct transformed predictions instead combine near-complete concept replacement with concepts whose fixed training-derived directions match the transformed motion at more than three times the 25\% random reference.
This analysis validates signed motion direction, where objective physical labels are available, rather than providing exhaustive semantic annotation or calibration for every fine-grained trajectory cluster.

\paragraph{Static-context stability.}
\label{appen:static_stability}
Using the same model and subset, we also test whether temporal reversal and rotation alter the semantics of the selected object and scene concepts.
For each concept group, we average the embedding vectors of the top-$k$ activated concepts ($k \in \{1, 3, 5, 10\}$) before and after perturbation and compute the cosine similarity between the two aggregate embeddings.
The similarity ranges from $0.84$ to $0.98$ under temporal reversal and from $0.81$ to $0.97$ under rotation, showing that motion perturbations alter the selected directional content while largely preserving the aggregate semantics of static context.

\subsection{Ground-Truth-Targeted Concept Interventions}
\label{appen:targeted_interventions}

The contribution-ranked removal results in \figref{deletion} and \tabref{faithfulness} show that \ours{} relies on high-contribution trajectory concepts.
However, contribution-ranked removal does not determine whether the influential concepts align with the ground-truth motion.
We therefore perform direction-targeted interventions on the Chiral SSv2 direction subset ($n=572$) and the TOMATO direction subset ($n=28$), using the fixed directional concept labels defined in \apref{directional_semantics}.
For each sample, we remove concepts from the ground-truth-direction, opposite-direction, or random non-ground-truth pool.
The random pool excludes concepts assigned to the ground-truth direction.

We consider two complementary interventions.
In count-matched removal, for sample $i$, let $G_i$ denote the set of ground-truth-direction concepts and $O_i$ the set of opposite-direction concepts.
We set $k_i=\min(\lvert G_i\rvert,\lvert O_i\rvert)$ and remove exactly $k_i$ concepts from the ground-truth-direction, opposite-direction, or non-ground-truth pool.
The non-ground-truth pool contains only opposite or orthogonal directions.
When a pool contains more than $k_i$ concepts, we randomly subsample $k_i$ concepts and average the result over $100$ draws.
Thus, all three conditions remove the same number of concepts for each sample, and the random control never includes ground-truth-direction concepts.
When $\lvert G_i\rvert>k_i$, some ground-truth-direction concepts remain active, making this a conservative test that controls intervention size without eliminating all same-direction redundancy.

Full-group removal instead deactivates every concept assigned to the ground-truth or opposite direction, testing the effect of removing potentially redundant concepts with the same directional semantics.
Its random non-ground-truth control removes $\lvert G_i\rvert$ concepts from the non-ground-truth pool and is averaged over $100$ draws.
The reported standard deviations therefore measure variability across concept-subsampling draws on the fixed evaluation set, not variability across training runs.
We report post-intervention accuracy and the correct-to-incorrect flip rate, the percentage of initially correct predictions that become incorrect after intervention.

\begin{table}[H]
\centering
\caption{\textbf{Ground-truth-targeted trajectory concept interventions.}
We report post-intervention accuracy, its change from the unablated model, and the flip rate, \ie the percentage of initially correct predictions that become incorrect.
Count-matched results and the random non-ground-truth full-group results report mean $\pm$ standard deviation over $100$ concept-subsampling draws on the fixed evaluation set.}
\label{tab:targeted_interventions}
\small
\setlength{\tabcolsep}{4pt}
\begin{tabular}{lllccc}
\toprule
Protocol & Task & Removed concepts & Acc. (\%) & $\Delta$Acc. & Flip rate (\%) \\
\midrule
\multirow{8}{*}{Count-matched}
& \multirow{4}{*}{AR} & None (\ours{}) & 81.1 & -- & -- \\
& & GT direction & $56.9{\pm}0.1$ & $-24.2$ & $30.1{\pm}0.1$ \\
& & Opposite direction & $82.2{\pm}0.1$ & $+1.1$ & $3.6{\pm}0.1$ \\
& & Random non-GT & $82.1{\pm}0.2$ & $+1.0$ & $3.7{\pm}0.2$ \\
\cmidrule(lr){2-6}
& \multirow{4}{*}{VQA} & None (\ours{}) & 46.4 & -- & -- \\
& & GT direction & $23.9{\pm}0.4$ & $-22.5$ & $70.5{\pm}0.5$ \\
& & Opposite direction & $54.0{\pm}0.2$ & $+7.6$ & $30.8{\pm}0.0$ \\
& & Random non-GT & $45.4{\pm}1.3$ & $-1.0$ & $21.3{\pm}2.0$ \\
\midrule
\multirow{8}{*}{Full-group}
& \multirow{4}{*}{AR} & None (\ours{}) & 81.1 & -- & -- \\
& & GT direction & 53.7 & $-27.4$ & 34.1 \\
& & Opposite direction & 81.3 & $+0.2$ & 4.5 \\
& & Random non-GT & $82.8{\pm}0.2$ & $+1.6$ & $3.8{\pm}0.1$ \\
\cmidrule(lr){2-6}
& \multirow{4}{*}{VQA} & None (\ours{}) & 46.4 & -- & -- \\
& & GT direction & 25.0 & $-21.4$ & 69.2 \\
& & Opposite direction & 53.6 & $+7.2$ & 30.8 \\
& & Random non-GT & $42.2{\pm}1.4$ & $-4.2$ & $25.4{\pm}2.1$ \\
\bottomrule
\end{tabular}
\end{table}

As shown in \tabref{targeted_interventions}, ground-truth-direction removal produces the largest degradation under both protocols.
On Chiral SSv2, count-matched removal reduces accuracy by $24.2$ points and changes $30.1\%$ of originally correct predictions to errors, whereas opposite-direction and random non-ground-truth removal do not reduce accuracy and yield flip rates below $4\%$.
Full-group removal produces the same ordering, reducing accuracy by $27.4$ points and yielding a $34.1\%$ correct-to-incorrect flip rate.
On TOMATO, ground-truth-direction removal reduces accuracy by $22.5$ points under count matching and $21.4$ points under full-group removal, with corresponding flip rates of $70.5\%$ and $69.2\%$.
Only $13$ TOMATO samples are predicted correctly before intervention, so these VQA flip rates should be interpreted descriptively.
A positive net accuracy change under a control can coexist with a nonzero correct-to-incorrect flip rate because overall accuracy also includes initially incorrect predictions that become correct.
Together with the directional analysis in \apref{directional_semantics}, these results show that trajectory concepts aligned with the ground-truth motion contribute disproportionately to both action recognition and VQA predictions.

\subsection{Video Question Answering Analysis}
\label{appen:native_vqa}

The VQA results in \tabref{main_results} show that \ours{} preserves competitive performance when predictions are routed through the typed trajectory/object/scene interface.
We further evaluate whether predictions on the native TOMATO and STAR benchmarks depend on trajectory concepts that contribute to the predicted answer.
For each sample, we rank trajectory concepts by their contribution to the concept-compatibility score for the predicted answer and set the selected trajectory concept activations to zero.
We remove concepts in descending (Top), random, or ascending (Bottom) order of contribution.

\begin{table}[H]
\centering
\caption{\textbf{Trajectory concept removal on native VQA benchmarks.}
For each sample, we remove the indicated fraction of trajectory concepts in descending (Top), random, or ascending (Bottom) order of contribution and report answer accuracy (\%).}
\label{tab:native_vqa_deletion}
\small
\setlength{\tabcolsep}{5pt}
\begin{tabular}{rcccccc}
\toprule
& \multicolumn{3}{c}{TOMATO} & \multicolumn{3}{c}{STAR} \\
\cmidrule(lr){2-4}\cmidrule(lr){5-7}
Removal ratio & Top & Random & Bottom & Top & Random & Bottom \\
\midrule
$0\%$   & 38.0 & 38.0 & 38.0 & 53.4 & 53.4 & 53.4 \\
$5\%$   & 26.1 & 33.7 & 38.0 & 12.9 & 52.1 & 53.4 \\
$10\%$  & 18.5 & 33.7 & 38.0 & 14.2 & 50.9 & 53.4 \\
$20\%$  & 16.3 & 33.7 & 38.0 & 11.7 & 48.8 & 53.4 \\
$40\%$  & 16.3 & 27.2 & 38.0 & 10.6 & 44.9 & 53.4 \\
$60\%$  & 19.6 & 30.4 & 38.0 & 11.5 & 40.3 & 53.4 \\
$80\%$  & 17.4 & 26.1 & 38.0 & 14.8 & 33.2 & 53.4 \\
$100\%$ & 23.9 & 23.9 & 23.9 & 24.7 & 24.7 & 24.7 \\
\bottomrule
\end{tabular}
\end{table}

As shown in \tabref{native_vqa_deletion}, removing only the top-$5\%$ trajectory concepts reduces accuracy by $11.9$ points on TOMATO and $40.5$ points on STAR.
The corresponding random removals reduce accuracy by $4.3$ and $1.3$ points, while Bottom-ranked removal has no effect through an $80\%$ removal ratio.
These results show that native VQA predictions depend on a sparse set of trajectory concepts with high contribution to the predicted answer rather than on arbitrary bottleneck dimensions.
At $100\%$ trajectory concept removal, object and scene concepts remain active, accounting for the nonzero residual accuracy.

The interpretable baselines evaluated in the main paper are designed for action recognition and do not directly define question-conditioned answer scoring.
We therefore construct two controlled variants that combine the MM-CBM~\cite{mm-cbm} concept-space matching design with language-derived motion concepts from LF-CBM~\cite{lf-cbm} or pose-sequence concepts from DANCE~\cite{dance}.
Because neither pose-sequence nor trajectory concepts have language names, the pose-sequence variant learns text-side relevance using the same answer-supervised alignment as \ours{}.
We additionally adapt SpLiCE~\cite{splice} to question-conditioned answer scoring.
All methods use the same frozen video and text backbones, data splits, answer candidates, and training budget.

\begin{table}[H]
\centering
\caption{\textbf{Comparison with VQA-capable concept methods.}
We report answer accuracy (\%) as mean $\pm$ sample standard deviation over five stochastic training seeds, except for the single-run SpLiCE adaptation.
The controlled variants combine the MM-CBM~\cite{mm-cbm} concept-space matching design with language-derived motion concepts from LF-CBM~\cite{lf-cbm} or pose-sequence concepts from DANCE~\cite{dance}.
The first three datasets are multiple-choice VQA reformulations of action-recognition datasets; TOMATO and STAR are native VQA benchmarks.}
\label{tab:vqa_concept_baselines}
\small
\setlength{\tabcolsep}{3pt}
\resizebox{\linewidth}{!}{%
\begin{tabular}{lcccccc}
\toprule
Method & NTU Mutual & Chiral SSv2 & UCF-101 & TOMATO & STAR & Avg. \\
\midrule
Non-interpretable baseline
& $97.4{\pm}0.1$ & $82.6{\pm}0.2$ & $99.8{\pm}0.0$ & $35.9{\pm}2.2$ & $54.5{\pm}0.4$ & 74.0 \\
Language-derived motion concepts
& $97.6{\pm}0.1$ & $82.0{\pm}0.4$ & $99.7{\pm}0.1$ & $31.7{\pm}3.6$ & $56.5{\pm}0.5$ & 73.5 \\
Pose-sequence concepts
& $97.6{\pm}0.1$ & $82.3{\pm}0.6$ & $99.7{\pm}0.1$ & $28.7{\pm}7.0$ & $55.9{\pm}0.7$ & 72.8 \\
SpLiCE~\cite{splice} adaptation
& 53.0 & 57.3 & 86.7 & 22.4 & 49.2 & 53.7 \\
\ours{}
& $97.1{\pm}0.3$ & $82.6{\pm}0.6$ & $99.7{\pm}0.1$ & $37.0{\pm}1.7$ & $55.4{\pm}1.2$ & \textbf{74.4} \\
\bottomrule
\end{tabular}%
}
\end{table}

As shown in \tabref{vqa_concept_baselines}, \ours{} achieves the highest mean accuracy across the five benchmarks among the concept-based methods ($74.4\%$), compared with $73.5\%$ for language-derived motion concepts and $72.8\%$ for pose-sequence concepts.
On the native VQA benchmarks, \ours{} obtains the highest TOMATO accuracy ($37.0\%$) and remains competitive on STAR ($55.4\%$).
We do not interpret the small numerical differences as significant improvements.
Together with the deletion results, these comparisons show that trajectory concepts support native VQA explanations while maintaining competitive predictive performance under the same concept-space matching design.

\paragraph{Concept combinations on STAR.}
\begin{table}[t]
    \centering
    \caption{\textbf{Effects of individual concept types on \ours{} performance.}}
    \vspace{0.2em}
    \resizebox{0.45\linewidth}{!}{ %
        \begin{tabular}{ccccc}
        \toprule
         \multicolumn{3}{c}{Concept} & \multirow{2}{*}{Acc.}  \\
         \cmidrule(lr){1-3} %
        Trajectory & Object & Scene &  \\
        \midrule
         \ding{51} &  \ding{55} & \ding{55} &  54.8 \\
         \ding{55} & \ding{51} & \ding{55} & 54.4\\
         \ding{55} & \ding{55} & \ding{51} & 52.9\\
         \ding{51}& \ding{51} & \ding{55} &   54.5\\
        \ding{51} & \ding{55} & \ding{51} & 52.9\\
        \ding{55} & \ding{51} & \ding{51} &   53.4 \\
        \ding{51} & \ding{51} & \ding{51} & 53.4\\
        \bottomrule
        \end{tabular}
    }
    \label{tab:vqa_concept_combination}

\vspace{-1.5em}
\end{table}

We evaluate combinations of trajectory, object, and scene concepts on STAR~\cite{star}.
We report single-run results for both STAR ablations in \tabref{vqa_concept_combination} and  \tabref{vqa_loss_ablation}.
As shown in \tabref{vqa_concept_combination}, trajectory concepts alone achieve $54.8$\% accuracy, compared with $54.4$\% for object concepts and $52.9$\% for scene concepts.
Combining all three concept types yields $53.4$\%.
Adding static context does not improve the reported accuracy over trajectory concepts alone in this evaluation.

\paragraph{Text-side trajectory relevance learning on STAR.}
\begin{table}
\centering
    \caption{\textbf{Effect of text-side trajectory relevance learning objectives.} }
\begin{tabular}{ l  c cc }
        \toprule
         Method &&& Acc. \\
        \midrule
         InfoNCE &&& 53.4\\
         Cosine similarity &&& 52.0 \\
        KL divergence &&& 52.6 \\
        $\ell_1$ regression &&& 52.1 \\
        $\ell_2$ regression &&& 52.2 \\
        \bottomrule
        \end{tabular}
        \label{tab:vqa_loss_ablation}
\end{table}

We compare loss functions for learning question-conditioned trajectory relevance on STAR~\cite{star}.
As shown in \tabref{vqa_loss_ablation}, InfoNCE achieves the highest accuracy ($53.4$\%), while the alternative objectives yield $52.0$--$52.6$\%.
The comparison supports our choice of InfoNCE for answer-supervised trajectory alignment.

\section{Limitations}
\label{appen:limitations}

While \ours{} provides effective and interpretable explanations by disentangling trajectory, object, and scene concepts, it has several limitations.
 
First, trajectory concepts depend on the quality of the underlying optical flow estimator.
Inaccurate flow predictions in challenging conditions---such as occlusion, fast motion, or low-texture regions---can produce noisy or misleading trajectories, which in turn degrade the quality of the discovered trajectory concepts.
Moreover, optical-flow trajectories can mix foreground entity motion with global camera motion.
In a diagnostic on $65$ Chiral SSv2 videos from six camera motion classes, \ours{} achieves $98.5\%$ accuracy, showing that trajectory concepts can represent task-relevant camera motion; however, this does not establish invariance to camera motion when it acts as a nuisance, and foreground masking (\tabref{foreground_gating}) is not equivalent to camera ego-motion compensation.
 
Second, due to the absence of explicit object and scene annotations in most video datasets, \ours{} relies on a Video-LLM to generate concept candidates and on a pretrained vision--language encoder to produce soft pseudo-labels.
Consequently, the resulting object and scene representations may inherit limitations of the underlying models, such as language priors and visual biases. 
Third, our concept visualizations rely on Grad-CAM~\cite{gradcam} to localize where a trajectory concept is expressed, so the highlighted regions are approximate. 
Fourth, trajectory concepts are extracted from short clips (\eg $L=16$ frames), which limits their ability to represent long-range temporal structure.
 
To mitigate sensitivity to optical flow quality and nuisance camera motion, future work could explore more robust flow estimators, video stabilization, global-motion subtraction, joint motion decomposition, or alternative trajectory representations that do not depend on dense flow fields.
To improve visualization precision, learned spatial grounding or attention-based localization could replace Grad-CAM~\cite{gradcam} for concept overlay.
Extending \ours{} to long-range temporal reasoning may involve generalizing trajectory concepts to broader visual dynamics, such as modeling temporal interactions among scene elements beyond pixel-level flow.

\section{Broader Impacts}
\label{appen:broader_impacts}

This work aims to advance explainability in video understanding by representing dynamic evidence as trajectory concepts and separating it from static object and scene context.
By deriving motion concepts from optical-flow trajectories, \ours{} enables source-flexible explanations that cover people, manipulated objects, and multi-agent interactions within a single framework.
This broader coverage makes trajectory-based explanations applicable to a wider range of real-world scenarios, such as object-centric manipulation and multi-agent environments, and may extend to egocentric video.
Furthermore, the same typed concept interface supports both action recognition and video question answering, demonstrating that concept-level explanations need not be redesigned for each task.
By exposing whether a model relies on meaningful motion cues, static contextual shortcuts, or spurious correlations, \ours{} can help practitioners diagnose and correct model errors through concept-level intervention, promoting trustworthy deployment in high-stakes domains such as healthcare, education, and human--AI interaction.
We hope this line of research fosters safer, more interpretable, and human-aligned video AI systems.
At the same time, explanations can be over-trusted: a plausible concept explanation does not by itself guarantee correct model behavior, so explanations should complement rather than replace rigorous evaluation. 
In addition, improved action recognition and interpretability tools could be misused for surveillance; we encourage their use in accordance with applicable privacy regulations and ethical guidelines.

\newpage

\end{document}